\documentclass[10pt,journal,compsoc]{IEEEtran}
\usepackage{graphicx}
\usepackage{comment}
\usepackage{amsmath,amssymb} 
\usepackage{color}
\usepackage{array}
\usepackage{url} 
\usepackage[hyperfootnotes=false]{hyperref}       
\usepackage{tabularx}
\newcolumntype{P}[1]{>{\centering\arraybackslash}p{#1}}
\usepackage{booktabs, multirow, array, makecell, caption}

\newcommand{\Revision}[1]{\textcolor{black}{{#1}}}

\ifCLASSOPTIONcompsoc
  \usepackage[nocompress]{cite}
\else
  \usepackage{cite}
\fi
\ifCLASSINFOpdf
\else
\fi
\begin{document}

\title{\Revision{Exposure Trajectory Recovery from Motion Blur}}

\author{Youjian~Zhang,
        Chaoyue~Wang,
        Stephen~J.~Maybank,~\IEEEmembership{Fellow,~IEEE,}
        and~Dacheng Tao,~\IEEEmembership{Fellow,~IEEE}
\IEEEcompsocitemizethanks{
\IEEEcompsocthanksitem Y. Zhang, C. Wang and D. Tao are with the School of Computer Science, Faculty of Engineering, University of Sydney, Darlington, NSW, Australia. \protect\\
E-mail: yzha0535@uni.sydney.edu.au, chaoyue.wang@sydney.edu.au, dacheng.tao@sydney.edu.au
\IEEEcompsocthanksitem S. Maybank is with the Department of Computer Science and Information Systems, Birkbeck College, Malet Street WC1E 7HX, London, UK. \protect\\
E-mail: sjmaybank@dcs.bbk.ac.uk
}
\thanks{Manuscript received Month Day, Year; revised Month Day, Year. \\
(Corresponding authors: Chaoyue Wang and Dacheng Tao.)}}

\IEEEtitleabstractindextext{%
\begin{abstract}
\Revision{Motion blur in dynamic scenes is an important yet challenging research topic. Recently, deep learning methods have achieved impressive performance for dynamic scene deblurring. However, the motion information contained in a blurry image has yet to be fully explored and accurately formulated because: (i) the ground truth of dynamic motion is difficult to obtain; (ii) the temporal ordering is destroyed during the exposure; and (iii) the motion estimation from a blurry image is highly ill-posed. By revisiting the principle of camera exposure, motion blur can be described by the relative motions of sharp content with respect to each exposed position. In this paper, we define exposure trajectories, which represent the motion information contained in a blurry image and explain the causes of motion blur. A novel motion offset estimation framework is proposed to model pixel-wise displacements of the latent sharp image at multiple timepoints. Under mild constraints, our method can recover dense, (non-)linear exposure trajectories, which significantly reduce temporal disorder and ill-posed problems. Finally, experiments demonstrate that the recovered exposure trajectories not only capture accurate and interpretable motion information from a blurry image, but also benefit motion-aware image deblurring and warping-based video extraction tasks. Codes are available on \url{https://github.com/yjzhang96/Motion-ETR}.} 

\end{abstract}

}

\maketitle

\IEEEdisplaynontitleabstractindextext

%
\IEEEpeerreviewmaketitle

\IEEEraisesectionheading{\section{Introduction}\label{sec:introduction}}

%
%
%
%

\IEEEPARstart{M}{otion} blur in the dynamic scene caused by camera shake, object motion, or depth variation is one of the commonest image degradations. Estimating motion information and restoring sharp content in dynamic blurry images would benefit many real-world applications including segmentation, detection, and recognition. Benefiting from the powerful fitting ability of deep convolutional neural networks (CNNs), deep learning-based deblurring methods~\cite{nah2017deep,gao2019dynamic,zhang2019deep,purohit2020region} have achieved impressive performance for dynamic motion blur removal. \Revision{Nevertheless, exploring motion information in a blurry image remains an academic and commercial challenge.}

Most conventional blur estimation/removal methods are based on blur kernel optimization~\cite{fergus2006removing,jia2007single,tai2010richardson,levin2011efficient,hyun2013dynamic, hyun2014segmentation,pan2019phase}, which assumes that a blurry area can be represented as a weighted sum of its latent sharp surrounding content. A blur kernel is a weighted matrix that performs convolution on a sharp image patch to synthesize a blurry pixel. Conversely, blur kernel estimation is cast as an energy minimization problem 
that aims to recover both the blur kernels and the latent sharp image from a blurry image. Such optimizations are highly ill-posed, so most conventional methods are restricted by assumptions of motion types and predefined image priors. For example, \cite{tai2010richardson,gupta2010single,whyte2012non,zheng2013forward} only handle blur caused by camera rotations, in-plane translations, or forward out-of-plane translations. For more complex dynamic motion blur, identifying a suitably informative and general prior is extremely difficult. 

Accompanying the development of deep neural networks, learning-based methods \cite{gong2017motion,sun2015learning} have been proposed to estimate blur kernels directly from blurry images. Compared to optimization-based methods, learning-based methods utilize predefined kernels to synthesize blurry data and then train an estimation network in a supervised manner. A well-trained estimation network is usually more effective and efficient at modeling object motion blur. However, due to the inherent limitations of blurry data synthesis, existing predefined blur kernels only cover limited motion types such as 2D vectors (\textit{i.e.,} linear motions), as in ~\cite{gong2017motion}. As a consequence, these methods may not be as effective for complex real-world dynamic scene blur.

Taking advantage of advanced photographic equipment, dynamic scene datasets \cite{nah2017deep,su2017deep} containing high frame-rate videos have been compiled to further understand dynamic motion blur. A real-world blurry image can be regarded as an accumulation of multiple ``instant'' frames, where a sequence of instant frames implicitly records blur (motion) information during an exposure period. Some methods \cite{jin2018learning,purohit2019bringing} are trained to directly recover these high frame-rate sharp frames (\textit{video extraction}) without explicitly depicting dynamic motions. Moreover, in some video deblurring studies~\cite{chen2018reblur2deblur,liu2020self}, optical flow is estimated between adjacent frames as another motion representation. However, since the optical flow between two frames is inherently linear and multiple frames may be misaligned, the estimated optical flow cannot perfectly match the dynamic motion contained in a single blurry image.

\Revision{Although recent years the end-to-end training models can directly recovery the sharp content \cite{tao2018scale,gao2019dynamic,zhang2019deep}, video sequence~\cite{jin2018learning,purohit2019bringing} or 3D scene~\cite{qiu2019world} from a blurry image, we argue that the motion information carried by a blurry image is important and has not been fully explored due to the aforementioned limitations, \textit{e.g.} synthetic ground truths, predefined priors, or temporal disorder. In this paper, our main target is to better estimate the motion information of a single dynamic blurry image without using any motion ground truth.}

\Revision{According to camera exposure principles, dynamic motion blur is caused by the relative motions of sharp content with respect to each exposed position. Inspired by this principle, we define the trajectories of these relative motions as \textit{exposure trajectories}. Compared to the convolutional blur kernel, the exposure trajectory can describe a specific physical movement in the temporal order.} \Revision{Through modeling pixel-wise displacements of the latent sharp image at continuous timepoints, exposure trajectories break the linear assumption in existing methods. In addition, we propose a novel motion offsets estimation framework to recover exposure trajectories of a single blurry image. Since the proposed differentiable motion offset module can easily be plug into CNNs for backpropagation, our motion/trajectory estimation framework is trained in a reblurring cycle that only need paired blurry/sharp image pairs. Moreover, to overcome the ill-posed nature of blur estimation and to model complex non-linear motions, we further apply a variety of constraints to ensure that the learned motion offsets form different types of trajectories, \textit{e.g.,} linear, bi-directional linear, or quadratic curves.}

\Revision{Besides recovering accurate and interpretable motion information from a blurry image, we successfully apply the recovered exposure trajectories to two downstream tasks}. For image deblurring, we devise a motion-aware deblurring module that takes pixel-wise trajectories to modulate the shape of convolution filters. Experiments show that the proposed motion-aware module enables a more effective deconvolution operation to handle large-scale dynamic motion blur with promising results. In addition, warping-based video extraction from a single blurry image can easily be achieved using the learned exposure trajectories. Compared to existing video extraction models, our solution generates the video with more accurate motions and is capable of interpolating an arbitrary number of middle frames, \textit{i.e.,} deriving slow-motion videos.

In summary, the contributions of this work are four-fold:

\begin{itemize} \vspace{-1 mm}
\setlength{\itemsep}{0pt}
\setlength{\parsep}{0pt}
\setlength{\parskip}{0pt}

\item \Revision{We propose a novel framework to model exposure trajectory, which represents the motion information contained in a dynamic blurry image. Compared with existing motion representations, \textit{e.g.,} conventional blur kernels, high-speed video frames (or estimated optical flow), and linear 2D motion vectors, our recovered exposure trajectories are more accurate and easier to interpret. Specifically, the causes of dynamic image blur are the first time modeled as dense, non-linear and continuous trajectories.}

\item \Revision{To recover exposure trajectories from a single blurry image, we proposed a motion offset estimation framework that contains a motion offsets estimation network and a blur creation module. To our best knowledge, it is the first differentiable image (re)blurring module that enables training an end-to-end motion estimation network without supervision on ground-truth motion information.}

\item \Revision{To address the ill-posed nature of dynamic motion/trajectory recovery, we propose multiple constraints such that the learned exposure trajectories follow certain patterns. We implement linear, bi-directional linear, and quadratic constraints, and in doing so demonstrate that our motion offsets with non-linear quadratic constraints outperform existing methods in fitting realistic dynamic motion blur.} 

\item \Revision{We present extensive experiments and analysis to demonstrate (i) our method can recover accurate and interpretable motion information from a single blurry image (Sec 5.3); (ii) the recovered exposure trajectories further benefit motion blur related tasks. By introducing recovered trajectories and motion-aware convolution, we improved the image deblurring performance over the baseline model (Sec 5.4). For extracting videos from a blurry image (Sec 5.5), though our model is only trained on blurry/sharp image pairs, it can synthesize high-quality videos with arbitrary frame rates and further delivers an impressive optical-flow field of the dynamic scene.}
\end{itemize}

\Revision{The rest of the paper is organized as follows. After a brief summary of the related works in Section~\ref{sec:relatedwork}, we illustrate our exposure trajectory recovery framework in Section~\ref{Sec:ETR}. Specifically, we first explain how to model the exposure trajectory, and then introduce the training scheme for exposure trajectory estimation. In Section~\ref{Sec:applications}, we attempt to apply the recovered exposure trajectories to image deblurring and video extraction tasks, which justifies the advantages of estimating exposure trajectories. Experiments in Section~\ref{Sec:Experiments} validate the accuracy of our trajectory estimation, improvements of deblurring performance, and superiority in video extraction, respectively. Limitations and failure cases are discussed in Section~\ref{Sec:limitations}. Finally, we conclude this paper with some future directions in Section~\ref{Sec:Conclusion}.}
	
\section{Related Work}\label{sec:relatedwork}

Single image blur estimation and removal have been extensively studied, with many methods proposed to solve different deblurring or blur estimation problems. Here, we focus our discussion on recent motion blur studies, reviewing optimization- and learning-based methods for blur estimation and removal, respectively.

\subsection{Optimization-based Methods}
	
A blur process is conventionally modeled as a convolution operation in which blur kernels are applied to a latent sharp image to generate a blurry output. Given a blurry image, optimization-based methods aim to iteratively recover its deblurred result and the blur kernels that model blur motions. However, this problem is ill-posed, so optimization-based methods adopt predefined image priors \cite{fergus2006removing,jia2007single,chen2019blind,shan2008high,cho2009fast,xu2010two,pan2016l_0,pan2017deblurring, li2021appealnet} or specific camera motion types \cite{nayar2004motion,gupta2010single,whyte2012non,zheng2013forward} to constrain the solution space of the blur kernels. For example, Tai \textit{et al.}~\cite{tai2010richardson} proposed a general projective motion model for cameras undergoing ego motion. Gupta \textit{et al.} \cite{gupta2010single} generalized camera motion to 2D translation and in-plane rotation and modeled them as motion density functions. Whyte \textit{et al.} \cite{whyte2012non,whyte2014deblurring} focused on solving the non-uniform blur caused by camera shake, aiming to recover the 3D rotation of the camera during an exposure process. Zheng \textit{et al.} \cite{zheng2013forward} attempted to handle another type of motion blur in which the camera moves primarily forwards or backwards by exploring homography associated with different 3D planes. Overall, under predefined priors/assumptions, the ill-posed optimization problem becomes solvable, and these methods have achieved reasonable performance on specific blurry data. However, most of these priors assume that the underlying scene is static and that the blur is caused by camera motion rather than the movement of objects in the captured image.

However, it is difficult to identify a suitably informative and general prior for object motion within a dynamic scene. Therefore, some authors \cite{pan2016soft,shi2014discriminative,hyun2013dynamic} have segmented different types of motion blur to overcome this problem. For example, Hyun \textit{et al.} \cite{hyun2013dynamic} proposed a novel energy function designed from the weighted sum of multiple blur data models. To handle different types of motion, their method estimated different motion blurs and their associated pixel-wise weights. Then,  \cite{pan2016soft} proposed soft-segmentation for object layer estimation. By jointly estimating object segmentation and camera motion, they achieved favorable object motion blur removal performance. Although motion segmentation seems an ideal extension of optimization-based methods, it is hard to estimate an accurate segmentation due to ambiguous pixels between regions. Furthermore, even within a segmented area, existing priors can only handle a limited number of motion types.
	
\subsection{Learning-based Methods}

In order to overcome the limitations of manually designed image priors or specific camera motions, learning-based methods aim to directly predict blur kernels (or deblurred results) from an input blurry image. Benefiting from the development of CNNs, learning-based models can be trained on a large amount of blurry data and can perform blur estimation (or removal) in an end-to-end manner.
	
Most learning-based methods were originally proposed to estimate blur causes/representations from blurry images~\cite{nayar2004motion,khare2011image,aizenberg2006blur,schuler2015learning}. For example, \cite{aizenberg2006blur,khare2011image} attempted to identify the type of blur  from a restricted set of parametrized blurs. Schuler \textit{et al.} \cite{schuler2015learning} proposed a CNN module for learning a gradient-like representation and estimated the blur kernels by dividing the learned representation in Fourier space. Similarly, \cite{chakrabarti2016neural} predicted the Fourier coefficients of a deconvolution kernel that modeled blind motions of an image patch. Sun \textit{et al.} \cite{sun2015learning} proposed a CNN-based model to predict the probabilistic distribution of motion blur at the patch level. In their method, a well-trained model estimated the direction and length of non-uniform linear motions. Then, \cite{gong2017motion} developed a fully convolutional framework to achieve pixel-wise prediction of blur kernels. Compared to optimization-based methods, these learning-based methods were more flexible and more efficiently estimated motion blur. However, during training, most learning-based methods required the ground truths of blur representations for supervision. Since the ground truths of real-world blurry data are rarely available, these methods were trained on artificially-generated training examples, limiting the approach to some simple blur types (\textit{e.g.,} linear motion). For more complex real-world dynamic motion, new blur representations and learning schemes are required to improve the estimations. 
	
Accompanying the increased fitting capability of CNNs, many learning-based methods have been proposed to directly restore the latent sharp image from a blurry input \cite{nah2017deep,tao2018scale,gao2019dynamic,kupyn2018deblurgan,zhang2018dynamic,nah2019recurrent,zhang2019deep,wang2019edvr,wang2019evolutionary}. Among these methods, \cite{nah2017deep} proposed a multi-scale network which performed deblurring in a ``coarse-to-fine'' pipeline. Then, \cite{tao2018scale,gao2019dynamic} further improved on this strategy by altering the parameter sharing and independent scheme. By combining three CNNs and a recurrent neural network (RNN), Zhang \textit{et al.} \cite{zhang2018dynamic} employed the learned variant RNN weights to model spatial-variant blurs. Inspired by \cite{zhang2018dynamic}, many methods \cite{suin2020spatially,purohit2020region,yuan2020efficient} have adopted a spatial-variant convolutional module as a substitute for some of the original convolution layers to increase the size of the receptive field in a more compact way. In addition, Kupyn \textit{et al.} \cite{kupyn2018deblurgan} and Ramakrishnan \textit{et al.} \cite{ramakrishnan2017deep} combined deblurring with generative adversarial networks (GANs) to synthesize more realistic sharp images. Overall, the combination of recently established real-world blurry datasets~\cite{nah2017deep} and the powerful learning capability of CNNs have allowed learning-based methods to achieve impressive performance for directly synthesizing deblurred images. Unfortunately, the causes of blur (motions) are generally ignored in these works, preventing the exploration of the rich dynamic information contained in blurry images and introducing training difficulties for related tasks due to a poor understanding of dynamic motion blur.  For example, in the absence of motion information, some deblurring and video extraction approaches either require a large receptive field to model large-scale blur~\cite{zhang2018dynamic} or require a complex training scheme and iterative inferences~\cite{jin2018learning,purohit2019bringing,zhang2020every}. In this work, we show that improving blur estimation can contribute to overcoming these problems and solving these tasks.

	\begin{figure*}[ht]
		\centering
		\includegraphics[width=\linewidth]{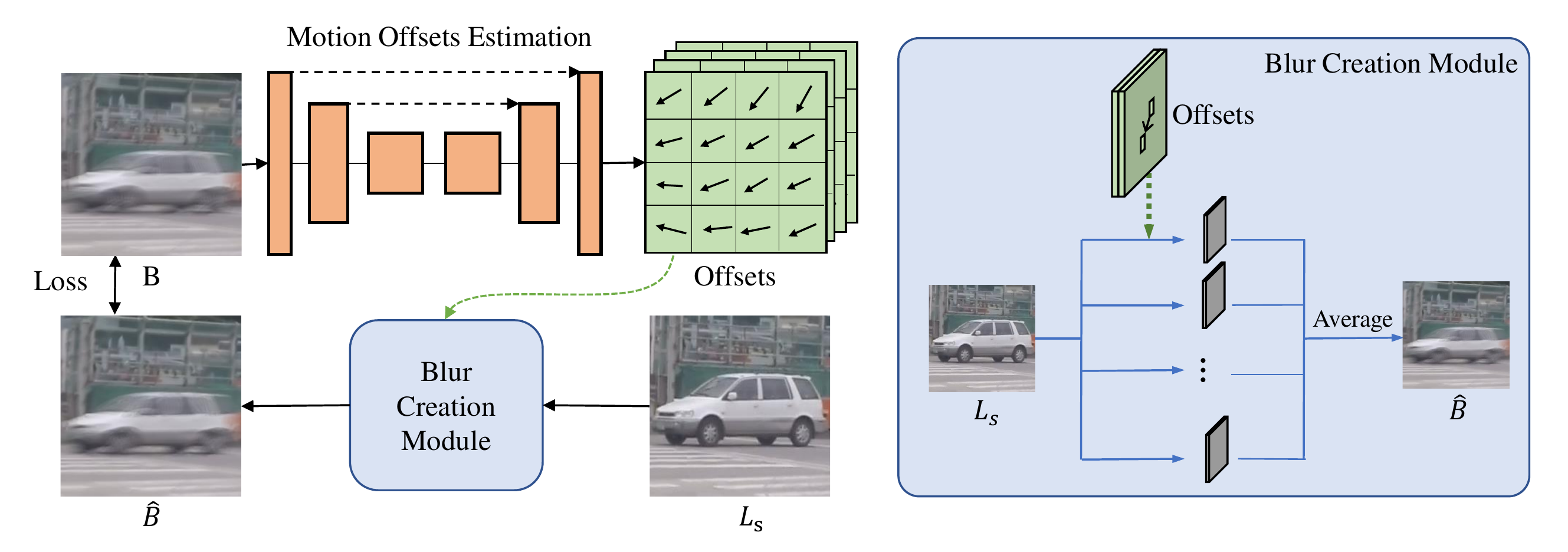} \vspace{-6 mm}
		\caption{Illustration of our proposed motion offset estimation method. The figure on the left is our motion offset generation network. It takes blurry images as input and outputs the corresponding motion offsets. Afterwards, the \textit{blur creation module} (on the right) takes a sharp image and the extracted motion offsets to reconstruct the input blurry image.}
		\label{Fig:offset estimation model}
	\end{figure*}

\section{\Revision{Exposure Trajectory Recovery}}\label{Sec:ETR}
\subsection{\Revision{Motion Exposure Mechanism}}\label{Sec:exposure model}
	
When a camera takes a photograph, the exposure time cannot be instant due to technological constraints and physics (\textit{i.e.,} exposure requirements). Therefore, a photograph records a target scene over a period of time. The exposure process can be formulated as:
\begin{equation} \label{eq:1} 
B = \int_0^\tau H(L,t) \ dt,
\end{equation}
where $L$ represents the latent content/scene in the photograph, $H(L,t)$ denotes the instant frame at time $t$, and $\tau$ denotes the camera exposure time. Due to camera shake or the motion or deformation of objects in the scene, $H(L,t)$ may continuously vary with respect to time, leading to dynamic scene blurry image $B$. 
    
In this work, we assume the middle instant sharp frame $L_s$ records all visual information of latent content/scene $L$.\footnote{\scriptsize{For most dynamic scene motion blur datasets, the middle instant frame is regarded as a sharp ground truth.}} According to the principle of camera exposure, the function $H(L_s, t)$ can be defined as an image wrapping operation that performs a pixel-wise shift over different times, \textit{i.e.},
\begin{equation}\label{eq:2}
H(L_s, t) = L_s(\textbf{P} + \Delta \textbf{P}^t),
\end{equation}
where \textbf{P} denotes all pixels in $L_s$, and $\Delta p^{t} = (\Delta x^t,\Delta y^t)$ is the shift of pixel $(x,y)$ at time $t$. 
Assuming the brightness remains constant during exposure, we consider Eq.~(\ref{eq:1}) and (\ref{eq:2}) and discretize them over multiple time steps $N$ to derive the formation of a blurry pixel $p_0$ as:
\begin{equation}\label{eq:5}
B(p_0) = \frac{1}{N} \sum_{n=0}^{N-1} L_s(p_{0} + \Delta p^{t_n}_{0}),
\end{equation}
which means a blurry pixel can be represented as the accumulation of pixels in the latent image moved by $\Delta p^{t_n}$. 

In this work, instead of deriving the blur kernels of a blurry image, we directly focus on the spatial shift $\Delta p^{t_n}$ of each pixel. Thus, we propose a new time-dependent blur representation, exposure trajectory which can be sampled by a set of motion offsets ($ {\{\Delta \textbf{P}^{t_n}\}}_{n=0}^{N-1} $). Similar to conventional blur kernels, the proposed exposure trajectory directly act on sharp images and then output blurry results. In contrast, our exposure trajectory models the blur formation as a spatial shift through time.
    
\subsection{Blur Creation Module}
\Revision{Based on the proposed exposure trajectory and motion exposure mechanism, we devise a \textit{blur creation module} which takes one sharp image $L_s$ and a set of motion offsets as inputs to generate a dynamic blurry image.} For each blurry pixel (\textit{i.e.,} exposure location) $p$, the proposed \textit{blur creation module} is asked to locate pixels $p^{t_n}=p+\Delta p^{t_n}$ in a latent sharp image $L_s$ (Eq.~(\ref{eq:5})) and further average them to obtain a blurry pixel. Since a real-world dynamic motion is continuous, we employ the bilinear interpolation to calculate the pixel value of location $p^{t_n}$,
\begin{equation}\label{eq:7}
L_s(p + \Delta p^{t_n}) = L_s(p^{t_n}) = \sum_{q}G(q,p^{t_n})\cdot L_s(q),
\end{equation}
where $q$ enumerates the referenced neighborhood points of the sampling location $p^{t_n}$, and $ G(\cdot,\cdot) $ is the bilinear interpolation kernel. As illustrated in Fig.~\ref{Fig:offset estimation model}, our motion offsets are of the same spatial resolution as the input image. Each offset has two channels corresponding to 2D axes. In practice, the \textit{blur creation module} takes $N$ motion offsets and a sharp image $L_s$ as inputs and synthesizes an averaged blurry output. 
    
\noindent\textbf{Discussion.}
Compared to conventional blur kernels, the proposed exposure trajectory (or motion offset) aims to mimic the exposure process of a camera sensor. If we assume the latent content/scene is known, motion offsets encode motion/dynamic information during an exposure period and can further synthesize a blurry image. Mathematically, the proposed motion offsets can be expressed in the formulation of blur kernels. Specifically, in the general blur kernel model, a blurry image $B$ is represented as $B = k \ast L + noise$, where $k$ represents a blur kernel. In such a framework, our motion offsets can be regarded as an equivalent blur kernel $k(p_0,t_n)$ of the location $p_0$ over time $\{t_n\}_{n=0}^{N-1}$,
\begin{equation}
k(p_0,t_n) = \begin{cases} \frac{\delta(p - (p_0 + \Delta p^{t_n}_0))}{N}, & \mbox{if $  p_0+\Delta p^{t_n}_0 \in L_s $} \\
0, & \mbox{otherwise} \end{cases}
\end{equation}
where $\delta (\cdot)$ denotes the Dirac delta function. 
	
Analyzing the equivalent formulation, the following differences between conventional blur kernels and our proposed motion offsets may exist. First, a time variable $t_n$ is introduced as a new and important element to model the exposure process. Under reasonable constraints (discussed in Section~\ref{Sec:different constraint}), our time-dependent motion offsets can act as a visualizable and interpretable representation of motion blur. 
Second, different from the weight matrix in a conventional blur kernel, our motion offsets calculate a uniform average of wrapped frames over a time sequence. We assume each time step is equally discretized; thus, the degree of motion (or blur) in each position is represented by the learned spatial displacements $\{ \Delta p^{t_n} \}$. Since the values of $\Delta p^{t_n}$ are continuous, bilinear interpolation (Eq.~(\ref{eq:7})) is performed to derive the value on each discrete pixel position. Note that the bilinear interpolation operation plays the same role as the weight matrix in blur kernels. Third, benefiting from the spatial shift operation (\textit{i.e.,} $\Delta p^{t_n}$), our equivalent blur kernel (\textit{i.e.,} motion offset) will not be limited by size, shape, pattern, or resolution, as traditional kernels are. For example, compared to the dense blur kernels estimated in~\cite{chen2018reblur2deblur}, where each kernel size is $33 \times 33$, the proposed motion offsets only carry $N\times 2$ parameters\footnote{As shown our experiments, setting $N$ as 15 already achieves extraordinary performance.}. Finally, since our motion offsets are compact and differentiable, the blur creation module can easily be integrated into deep neural networks and trained in an end-to-end manner.
	
    \begin{figure*}[ht]
        \centering
        \includegraphics[width=0.95\linewidth]{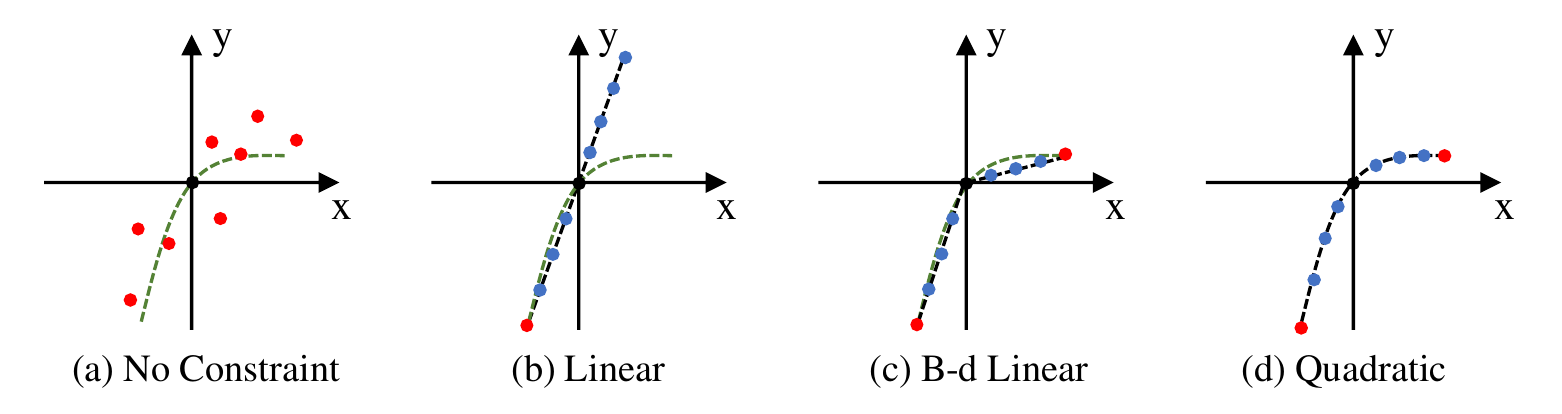}
        \vspace{-3mm}
        \caption{Examples of motion offsets with different constraints. Suppose the green curve is the ground truth exposure trajectory. (a)-(d) simulate the fitting results of motion offsets with no constraint, linear constraint, b-d linear constraint, and quadratic constraint, respectively. Red points are the offsets that output by the estimation network and blue points are calculated by the different constraints.}
        \label{fig:trajectory}
    \end{figure*}

\subsection{Motion Offset Estimation}\label{Sec:offset estimation}

	
\Revision{The biggest problem of previous learning-based blur kernel (motion) estimation is the universal absence of ground truth blur kernels for real-world data. Thus, \cite{gong2017motion,sun2015learning} must use synthetic data for training. Based on the motion exposure mechanism, the proposed motion offset could replace conventional blur kernels. Exploiting its compact and differentiable advantages, we devise a  training scheme that performs motion offset estimation without any ground truths of the motion information. Specifically, the \textit{blur creation module} is connected with the motion offset estimation network to form a cyclical pipeline.} 

Given a ground truth blurry image $B$ and a sharp reference frame $L_s$, the motion offset estimation network takes $B$ as an input and outputs $N$ motion offsets; then, the blur creation module takes $L_s$ and the motion offsets as input to reproduce the estimated blurry image $\hat{B}$. Fig. \ref{Fig:offset estimation model} illustrates this procedure. The motion offset estimation network is based on an encoder-decoder network with skip connections, and the detailed model structure is provided in Section~\ref{sec:implement}. 
    
The loss of this cyclic reconstruction can be written as:
\begin{equation}
\mathcal{L}_{circle} = \mathcal{L}_{l_2} + \lambda_{SSIM} \mathcal{L}_{SSIM},
\end{equation}
where $\mathcal{L}_{l_2}$ and $\mathcal{L}_{SSIM}$ denote the $\ell_2$ loss and \textbf{SSIM} loss respectively. Both are applied to measure the difference between $B$ and $\hat{B}$. We elaborate these two terms as follows:
\begin{align}
\mathcal{L}_{2} &= || B - \hat{B} ||_2^2, \\
\mathcal{L}_{SSIM}(P) &= 1 - \textbf{MS-SSIM}(\tilde{p}),
\end{align}
where $ \tilde{p} $ is the center pixel of patch $P$, and \textbf{MS-SSIM} denotes the multi-scale SSIM. A more specific definition and implementation can be found in \cite{zhao2016loss}. The reason that we use the SSIM loss is that the $\ell_2$ loss only weakly penalizes our output because it tends to generate average results, and the blurry image is already averaged. In this case, the SSIM loss more accurately measures the distance between two blurry images.  
	
We also introduce other losses to regularize motion offsets. First, we apply a regularization loss to encourage offsets that search for nearby pixels as solutions. This benefits the situation in which there is a large smooth region, \textit{e.g.,} the sky or ground, where large displacements (offsets) should be suppressed. Moreover, due to dynamic motion blur usually being continuous along the space, we apply the total variation loss to encourage spatial smoothness within offset maps. These two losses can be formulated as:

\begin{equation}
\mathcal{L}_{reg} = \frac{1}{Nwh}\sum_{n=1}^{N}\sum_{i=1}^{w}\sum_{j=1}^{h} M_{n}(i,j)^2, \\
\end{equation}

\begin{footnotesize}
\begin{equation}
\begin{split}
\mathcal{L}_{tv} = \frac{1}{N}\sum_{n=1}^{N} \Big(  &\frac{1}{(w-1)h}\sum_{i=0}^{w-1}|M_n(i,j)-M_n(i+1,j)|	+ \\ &\frac{1}{w(h-1)}\sum_{j=0}^{h-1}|M_n(i,j) - M_n(i,j+1)| \Big),
\end{split}
\end{equation}
\end{footnotesize}
where $M_{n}(i,j)$ denotes the location $(i,j)$ in the $n^{th}$ offset map.
	
In summary, the final loss function is a weighted sum of the above losses:
\begin{equation}
\mathcal{L} = \mathcal{L}_{circle} + \lambda_{reg}\mathcal{L}_{reg} + \lambda_{tv}\mathcal{L}_{tv}.
\end{equation}

\subsection{Different Constraints to Motion Offsets}\label{Sec:different constraint}
    
If we directly learn all the motion offsets using the framework described above, namely a zero constraint (ZC) model, the results will be as shown in Fig.~\ref{fig:trajectory}~(a). Though achieving impressive reblurring accuracy, its ill-posed nature creates the following problems: (1) the learned motion offsets are one of several possible solutions of blur formation, and since it is difficult to form them into an explicit trajectory as in real-world blur formation, the learned motion offsets are usually sub-optimal for describing realistic motion; and (2) although there exists the temporal variable $t_{n}$ in our learned offsets, these offsets are disordered due to a lack of spatial-temporal relationship modeling.
	
Therefore, we devise several constraints to reduce the ill-posed nature of motion estimation and to form the motion offsets into an explainable exposure trajectory:

\noindent\textbf{(Bidirectional) linear trajectory constraint.} 
The linear assumption is used to fit motion blur in many methods \cite{hyun2014segmentation,sun2015learning,gong2017motion}. We also devise a linear trajectory constraint for motion offsets. Recall our assumption (Section~\ref{Sec:exposure model}) that the sharp image $L_s$ represents the middle instant frame, \textit{i.e.} $\Delta p^{t_\text{mid}}=(0,0)$. To represent linear motion, the motion offset estimation network only needs to predict another point on the exposure trajectory. Suppose the blurred pixel is caused by uniform linear motion and the predicted offset $ \Delta p$ is an endpoint of the exposure trajectory, the other offsets can be derived as:
    
\begin{equation}
    \Delta p^{t_n} = (1 - \frac{2n}{N-1})\Delta p, n=0,\dots,N-1.
\end{equation}
We attempt to predict the furthest point (endpoint) of the exposure trajectory based on the observation that the blurred edge is easier to capture and estimate.
    
Taking a further step, we can apply a bidirectional linear (b-d linear) constraint to our motion offsets. As shown in Fig. \ref{fig:trajectory}~(c), we predict two offsets $\Delta p_1$, $\Delta p_2$ to represent the start and end points of each exposure trajectory. Then, the other offsets can be calculated as: 
\begin{equation}
\Delta p^{t_n} = \begin{cases} (1-\frac{2n}{N-1})\Delta p_1, & n = 0,\dots,\frac{N-1}{2}, \\
(\frac{2n}{N-1} - 1)\Delta p_2. & n = \frac{N+1}{2},\dots, N-1.
\end{cases}
\end{equation}
As shown in Fig.~\ref{fig:trajectory}, this trajectory better fits a curve than the linear one.

\noindent \textbf{Quadratic trajectory constraint.}
Although the bi-directional linear constraint already introduces a certain non-linearity into trajectory learning, the quadratic function can better approximate real-world motion~\cite{ren2017video,xu2019quadratic}. A quadratic curve can be derived when an object is moving with constant acceleration, a much stronger fitting than the (bi-)linear assumption. Thus, we devise a quadratic trajectory constraint to force a smooth quadratic trajectory on our motion offsets. Unlike previous works, which apply a quadratic trajectory between video frames, we extract this trajectory inside a single blurry frame. Specifically, we still predict two offsets $\Delta p_1$, $\Delta p_2$ as the start and end points of the exposure trajectory, with the other offsets written as:
\begin{equation}
\begin{aligned}
\Delta p^{t_n} & = \frac{\Delta p_1 + \Delta p_2}{2}(\frac{2n}{N-1}-1)^{2} \\
& + \frac{\Delta p_2 - \Delta p_1}{2}(\frac{2n}{N-1}-1), n=0,\dots,N-1.
\end{aligned}
\end{equation}
Thus, motion offsets will be formed into a quadratic trajectory (Fig. \ref{fig:trajectory}~(d)). Note that since our motion offsets are modeled in equidistant time, the learned motion offsets not only match a curvilinear exposure trajectory but also reflect the changing velocity. For example, a longer displacement between adjacent time steps corresponds to faster movement.

\section{\Revision{Applications Benefiting from Recovered Exposure Trajectories}}\label{Sec:applications}
\Revision{In this section, we apply our recovered exposure trajectory to two motion blur-related downstream tasks, \textit{i.e.} image deblurring and video extraction from a single blurry image, which further demonstrates the benefit of recovering accurate motions from a blurry image.}

\subsection{Motion-aware Image Deblurring}
    
\begin{figure*}[t!]
    \centering
    \includegraphics[width=0.95\textwidth]{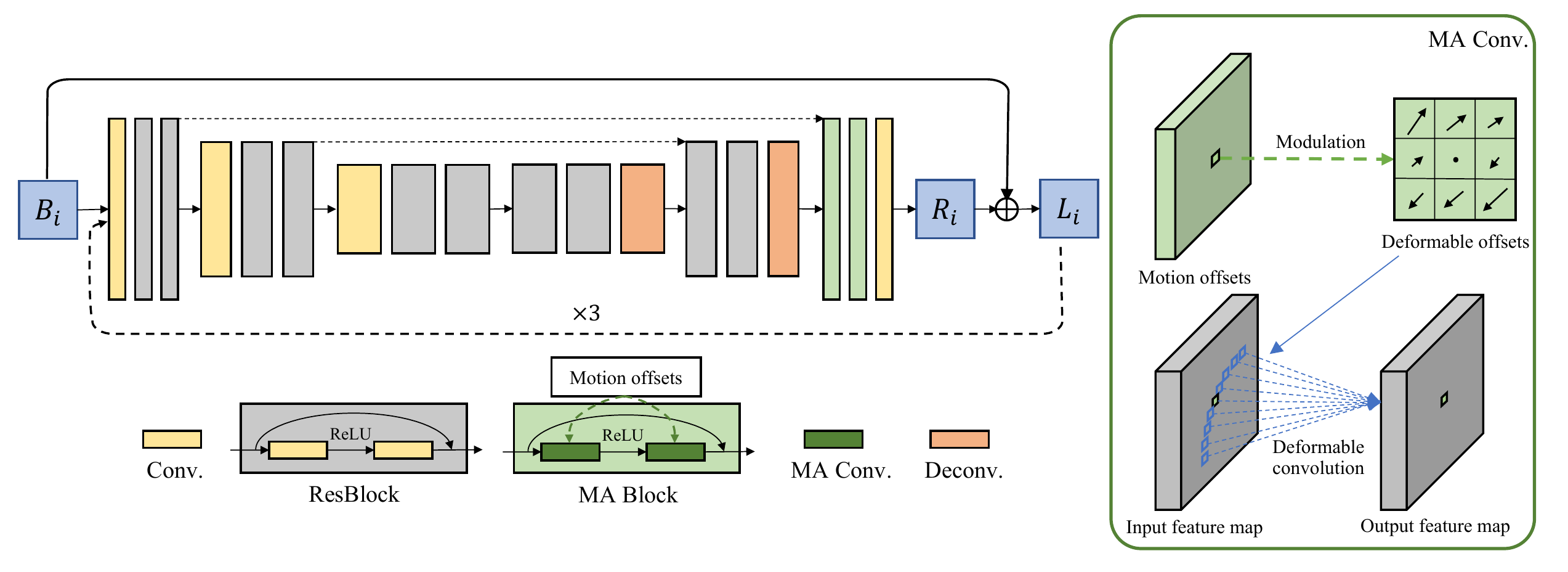}
    \vspace{-2mm}
    \caption{The proposed motion-aware deblurring network. An encoder-decoder residual architecture for image deblurring is shown on the left, while the schematic of a motion-aware convolution in the motion-aware block is shown on the right.}
    \label{fig:deblur network}
\end{figure*}
    
To handle the challenging problem of dynamic scene deblurring, existing works employ complex network architectures to enlarge the model capacity, such as a multi-scale structure~\cite{nah2017deep,tao2018scale,gao2019dynamic}. Some other methods~\cite{zhang2018dynamic,purohit2020region,suin2020spatially} claim that spatially invariant convolution filters, \textit{i.e.} spatially uniform and limited receptive fields are sub-optimal for modeling dynamic scene blur. With the recovered exposure trajectories, we aim to design a spatial-variant deblurring network, which leads to a more compact and efficient model.


\Revision{As derived in \cite{zhang2018dynamic,purohit2020region}, the blurry image deconvolution can be written as an Infinite Impulse Response (IIR) formula, from which they drew two main conclusions: 1) the deblurring process requires a very large receptive field; and 2) for a CNN-based deblurring model, the convolution filters should have a similar direction/shape with the blur kernel}. For example, if a blur kernel is linear and horizontal, the latent pixels can be calculated using only horizontal blurry pixels, thus the deconvolution filters should also be pure horizontal. 
\Revision{However, the conventional square-shaped convolutional filter cannot meet these requirements.} In this work, we propose a motion-aware deblurring network with spatial-variant convolution filters that are shaped by the recovered exposure trajectories.
    
To build a spatial-variant convolution module, the deformable convolution unit~\cite{dai2017deformable} provides a general solution. In recent works~\cite{purohit2020region,suin2020spatially}, spatial-variant deblurring modules based on deformable convolutions have achieved reasonable performance. However, since the ground truth of the kernel shape is absent, these methods attempt to derive deformation offsets from encoded features of an input blurry image. We propose the motion-aware convolution (MA Conv.), which directly employs the recovered motion offsets to model the aforementioned filter deformation. Our motion-aware convolution can be formulated as:
\begin{equation}
    y(p_{0}) = \sum_{n=0}^{N} w(p_{n}) \cdot x(p_{0} + \alpha \Delta p_{0}^{n}),
\end{equation}
where $x$ is an input feature map, $y$ is an output feature map, and $w$ is the weight of the convolution filter. \Revision{The coordinate $p_{0} + \alpha \Delta p_{0}^{n}$ denotes the sampling location calculated by the centering coordinate $p_{0}$ and an offset $\alpha \Delta p_{0}^{n}$, which controls the shape and size of the convolution, and $w(p_{n})$ is the weight corresponding for the sampling point $p_{0} + \alpha \Delta p_{0}^{n}$. For a square-shaped $3 \times 3$ convolutional filter with dilation 1, $\Delta p_{0}^{t_n} \in {(-1,-1),(-1,0),\dots,(0,1),(1,1)}$ and $\alpha=1$. In our motion-aware convolution, $\Delta p_{0}^{n}$ are decided by our recovered motion offsets. Specifically, giving the recovered exposure trajectory, we calculate 9 $\Delta p_{0}^{n}$ centered by the middle offset $\Delta p_{0}^{\text{mid}}=(0,0)$ to modulate the original $3 \times 3$ kernel, as shown in Fig.~\ref{fig:deblur network}. Moreover, we use hyper-parameter $\alpha$ to scale the size of the convolution, and $\alpha$ is set to 0.1 in our experiments.}
In this way, the proposed motion-aware convolution takes full advantage of the information contained in motion offsets, \textit{i.e.} both direction and magnitude, resulting in a more mathematically accurate deconvolution.
    

Here, we adopts the DMPHN(1-2-4)~\cite{zhang2019deep} as the backbone architecture of our motion-aware deblurring network, since it is relatively compact among the state-of-the-art models. As shown in Fig.~\ref{fig:deblur network}, similar with the most existing image deblurring methods, the encoder-decoder structure is employed. Compare to the vanilla DMPHN(1-2-4), our motion-aware deblurring network can be easily derived by replacing the selected convolutional layers with the proposed motion-aware convolution.
According to our experiments, adding the motion-aware convolutions in the last stage of the decoder achieved the best performance. In addition, to build a compact deblurring network, we do not employ the stack-DMPHN as Zhang \emph{et al} \cite{zhang2019deep}. Adding the motion-aware module can already achieve comparable results, and our model largely reduces the memory cost. 


\subsection{Warping-based Video Extraction from a Single Blurry Image}

Different from conventional blur kernels, our motion offsets contain temporal information that could help us to restore time series from a blurry input. As indicated in Eq.~(\ref{eq:2}), frame $L^{t_n}$ can be obtained through a transformation $H(\cdot,\cdot)$. Now, with the deblurring result $\hat{L_s}$ and the estimated motion offsets $\hat{\textbf{P}}^{t_n}$, we can generate the estimated frame $\hat{L}^{t_n}$:
\begin{equation}\label{eq:15}
\hat{L}^{t_n} = \hat{L_s}(\textbf{P} + \Delta \hat{\textbf{P}}^{t_n}).
\end{equation}
\Revision{Unlike forward optical flow which is a many-to-one mapping, our motion offset is equivalent to backward warping flow. Specifically, the backward warping flow represents the pixel displacement from warped result to the warping input. Therefore, there will be no holes in our warped result.} According to Section \ref{Sec:different constraint}, since we have added different trajectory constraints, theoretically we can interpolate arbitrary $N$ offsets into our start and end offsets, which further leads to smooth or even slow-motion video output. 
	
To our best knowledge, only two existing works have been capable of restoring a video sequence from a single blurry image. \cite{jin2018learning} first attempted to generated a video sequence from a single blurry image by training different networks to generate frames at different time $t_n$, only producing limited frames. \cite{purohit2019bringing} proposed a recurrent network to address temporal ambiguity, inferring the recurrent state at each time step $t_n$. Unlike these methods, we only need to calculate our motion offsets once, which is more time efficient. Moreover, these previous methods are trained with a series of ground truth sharp frames for supervision, which limit the generated outputs to specific time intervals. Our motion offset estimation module is easy to train and requires fewer annotations. Moreover, during test, our model is more compact and efficient. 

\begin{table}[t!]
    \centering
    \small
    \setcellgapes{1pt}\makegapedcells
    \caption{Detailed architecture of the motion offset estimation network. $+$ denotes that a skip connection concatenates this layer with the corresponding layer in the encoder.}
    \begin{tabular}{c|c|c}
        \hline
        Stage  & Output & Layer Details \\
        \hline
               & $\frac{H}{2} \times \frac{W}{2}$ & Space to Depth \\
        \hline
        Conv1 & $\frac{H}{2}\times \frac{W}{2}$ & $5\times5$, 12, 16, stride 1 \\
        \hline
        ResBlock1 & $\frac{H}{2}\times \frac{W}{2}$ & $\left[\begin{array}{l} 5\times5,16 \\5\times5,16\end{array}\right]\times 3$ \\
        \hline
        Conv2 & $\frac{H}{4} \times \frac{W}{4}$ & $5\times5$, 16, 32, stride 2\\
        \hline
        ResBlock2 & $\frac{H}{4}\times \frac{W}{4}$ & $\left[\begin{array}{l} 5\times5,32 \\5\times5,32\end{array}\right]\times 3$ \\
        \hline
         Conv3 & $\frac{H}{8} \times \frac{W}{8}$ & $5\times5$, 32, 64, stride 2\\
       \hline
        ResBlock3 & $\frac{H}{8}\times \frac{W}{8}$ & $\left[\begin{array}{l} 5\times5,64 \\5\times5,64\end{array}\right]\times 3$ \\
        \hline
        Bottleneck1 & $\frac{H}{8}\times \frac{W}{8}$ & $\left[\begin{array}{l} 1\times1, 64, 128 \\ 3\times3, 128, 64\end{array} \right]$ \\
        \hline
         Dconv1 & $\frac{H}{4} \times \frac{W}{4}$ & $5\times5$, 64, 32, stride 2\\
           \hline 
        Bottleneck2 & $\frac{H}{4}\times \frac{W}{4}$ & $\left[\begin{array}{l} 1\times1, 32+32, 128 \\ 3\times3, 128, 64\end{array} \right]$ \\
        \hline
         Dconv2 & $\frac{H}{2} \times \frac{W}{2}$ & $5\times5$, 64, 16, stride 2\\
       \hline 
        Bottleneck3 & $\frac{H}{2}\times \frac{W}{2}$ & $\left[\begin{array}{l} 1\times1, 16+16, 64 \\ 3\times3, 64, 32\end{array} \right]$ \\
        \hline
         Dconv3 & $H \times W$ & $5\times5$, 32, 32, stride 2\\
        \hline
        Conv4 & $H \times W$ & $5\times5$, 32, 4, stride 1 \\
        \hline
    \end{tabular}
   
    \label{tab:offset nestwork}
\end{table}

\begin{figure*}[ht!]
    \centering
    \includegraphics[width=0.95\linewidth]{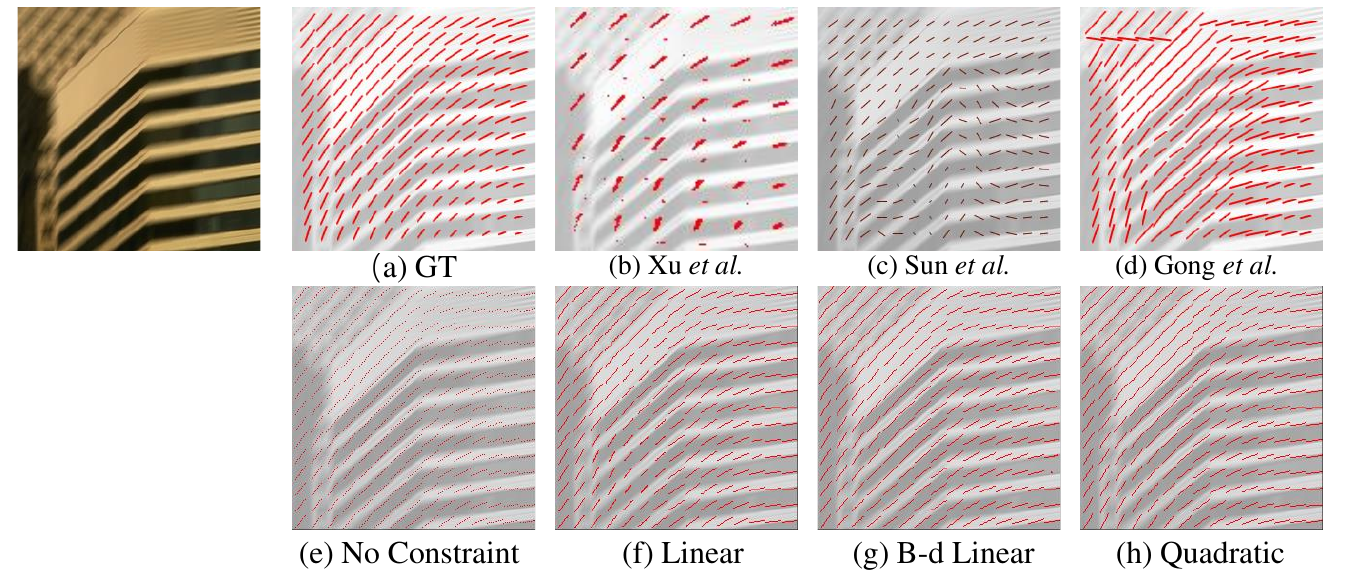}
     \vspace{-3mm}
    \caption{Examples of motion estimation on the synthetic dataset. The top row shows the blurry input, ground truth motion, and results of previous methods. The bottom row shows our estimated motion offsets under different constraints.}
    \label{fig:motion_syn}
\end{figure*}

\section{Experiments}\label{Sec:Experiments}

In this section, we first introduce our training configuration before carrying out quantitative and qualitative comparisons between our method and state-of-the-art methods for motion estimation, image deblurring, and video extraction.

\subsection{Implementation Details}\label{sec:implement}
We provide layer-wise details of our motion offset estimation networks in Table \ref{tab:offset nestwork}. $H$ and $W$ represent the height and width of an input blurry image. For training both the \textit{motion estimation network} and \textit{deblurring network}, we use Adam \cite{kingma2014adam} for optimization, with $ \beta_{1}=0.9 $, $\beta_{2}=0.999$ and $\epsilon=10^{-8}$. The learning rate is set initially to $10^{-4}$ and it is linearly decayed to 0. For motion offset estimation, we set the offset number to $N=15$, $\lambda_{SSIM}=0.1$,  $\lambda_{reg}=0.00002$, $\lambda_{tv}=0.0005$. All weights are initialized using Xavier \cite{glorot2010understanding}, and bias is initialized to 0. \Revision{We first train the motion estimation network using the blurry image and the ground-truth sharp image as the training pair. Then we train the deblurring network with the pre-trained motion estimation network. To train the deblurring network, besides paired training images, the estimated exposure trajectories are utilized as an auxiliary input.}

\begin{figure*}[ht!]
    \centering
    \includegraphics[width=0.97\linewidth]{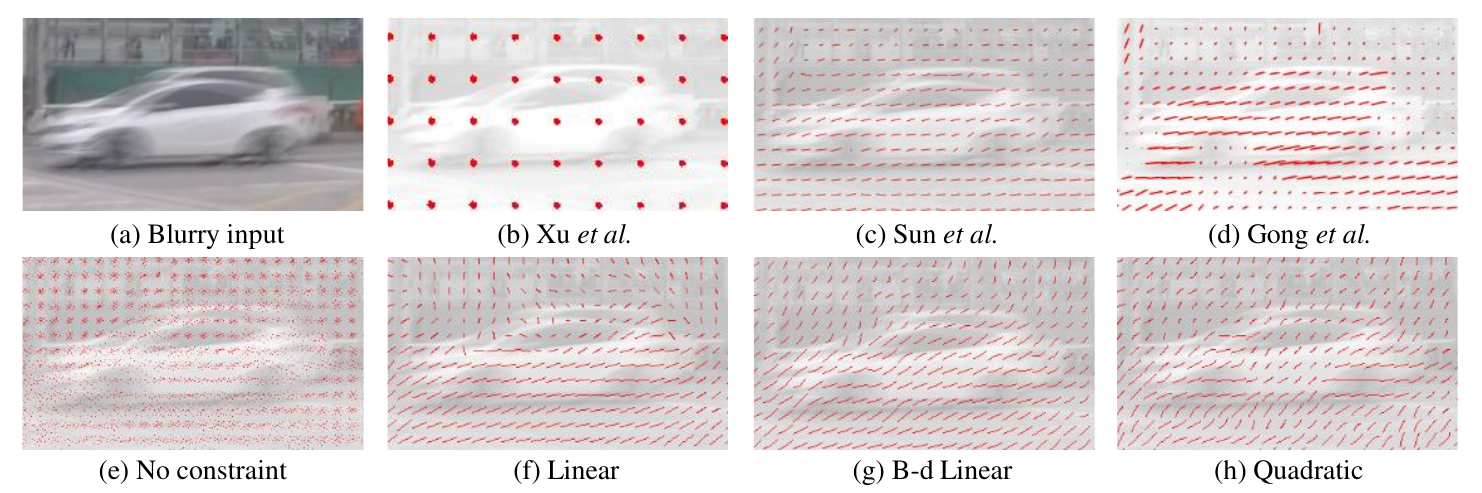}
    \vspace{-2.5mm}
    \caption{Examples of motion estimation on the GoPro dataset. The top row shows the blurry input and results of previous methods. The bottom row shows our estimated motion offsets under different constraints.}
    \label{fig:motion_gopro}
\end{figure*}

\begin{figure*}[ht!]
    \centering
    \includegraphics[width=0.95\linewidth]{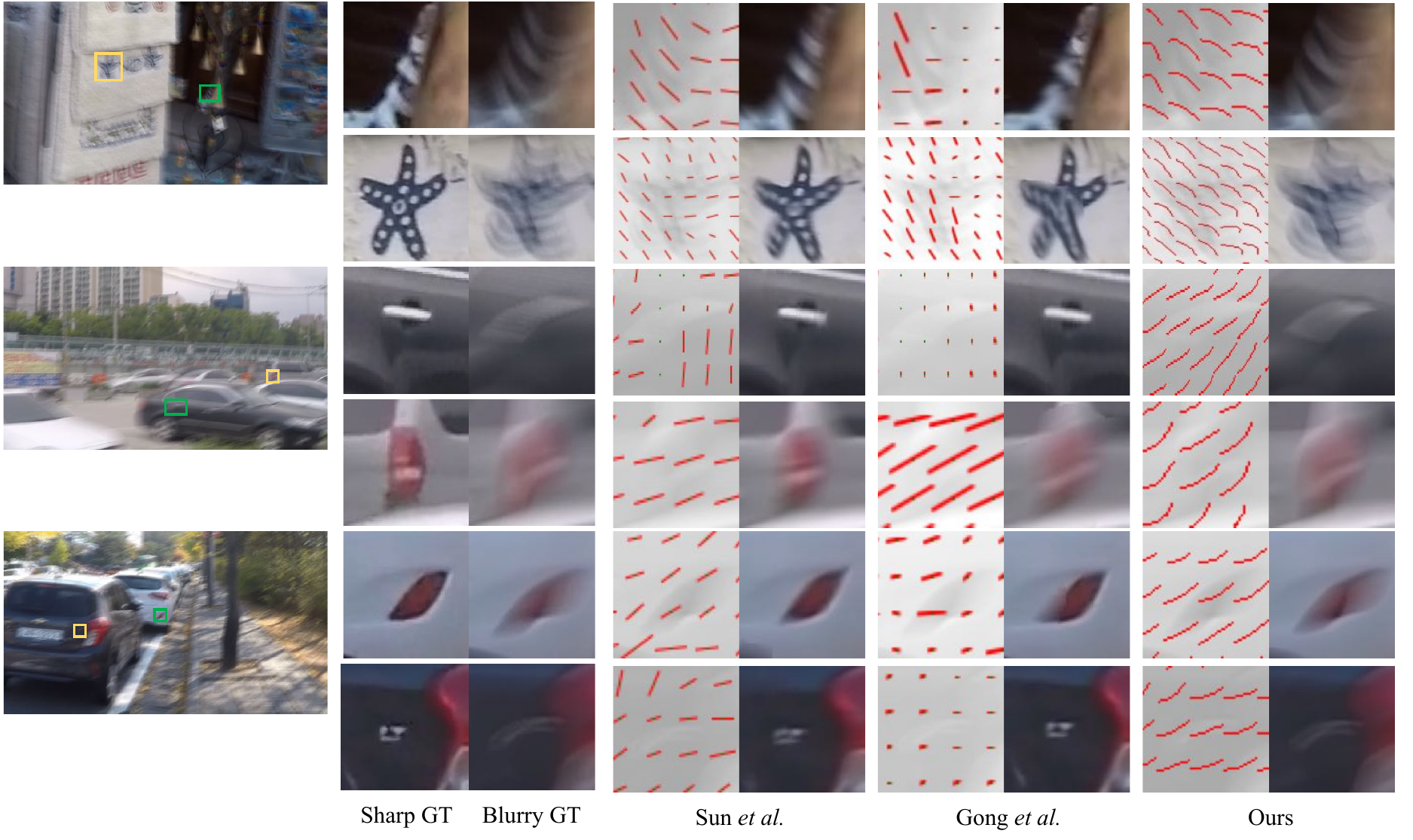}
    \vspace{-3mm}
    \caption{\Revision{Visual comparison of motion estimation on GoPro test set. From left to right shows the groud-truth image patches, the results of Sun \textit{et al.} \cite{sun2015learning}, Gong \textit{et al.}~\cite{gong2017motion} and our quadratic model. For each method, we visualize the estimated motion field and the reblurred result.}}
    \label{fig:motion_gopro_reblur}
\end{figure*}

\subsection{Datasets}
We employ two different datasets. The synthetic dataset provides ground truth blur kernels, while the GoPro dataset is synthesized from real-world frame with more challenging dynamic motion blur without ground truth blur kernels.

\textbf{Synthetic Dataset.}	We follow the same approach as in \cite{gong2017motion} to generate blurry/sharp image pairs with pre-defined blur kernels. Specifically, blur kernels are represented by a motion flow map filled with pixel-wise non-uniform motion vectors. Each vector can form a linear blur kernel. Same as \cite{gong2017motion}, we use images from BSD500 \cite{arbelaez2010contour}, which consists of 200 training images and 100 test images, as sharp ground truths. We then generate 50 motion flow maps for each training image and 3 motion flow maps for the test images. Finally, the sharp images are convolved with the corresponding flow maps to generate blurry images.

The \textbf{GoPro Dataset} \cite{nah2017deep} addresses the problem that synthetic data are different from real-world blurry images containing more complex dynamic motion. More realistic blurry images are generated by averaging consecutive short-exposure frames from a high frame rate video, \textit{e.g.,} 240fps, taken from a GoPro camera. In this way, \cite{nah2017deep} collected 3214 blurry/sharp image pairs, and split them into a training set with 2103 pairs and a test set with 1111 pairs. In following experiments, unless stated, the quantitative results are based on the GoPro dataset.

\begin{figure*}[ht!]
    \centering
    \includegraphics[width=\linewidth]{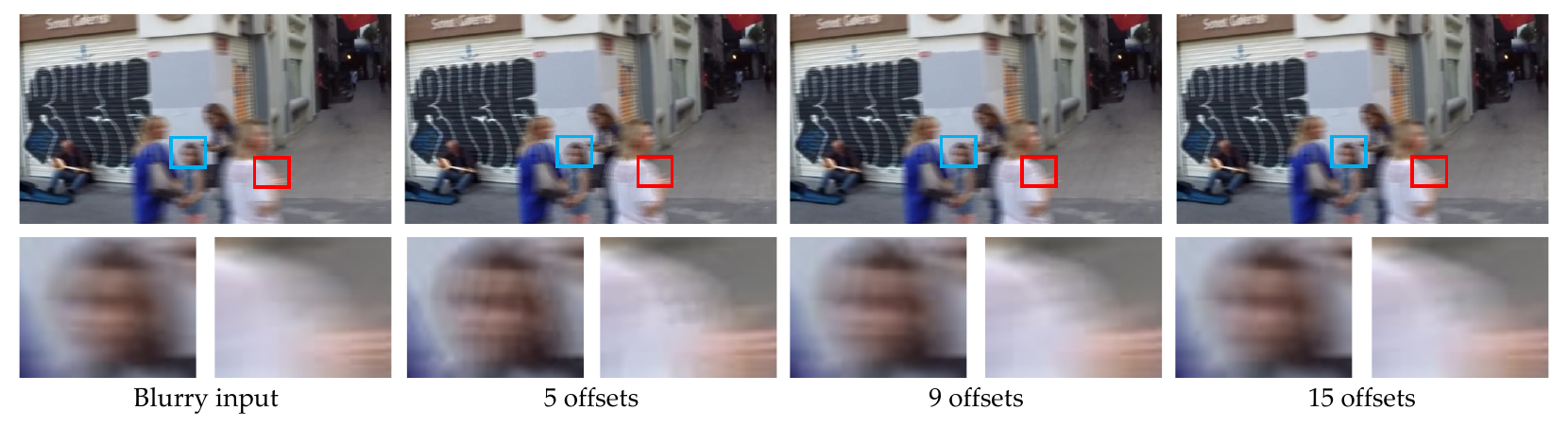}
    \vspace{-7 mm}
    \caption{The effect of offset number on blur creation. Left to right show the ground truth blurry image, the result of the model with 5 offsets, the result of the model with 9 offsets, and the result of the model with 15 offsets. It is clear that increasing the number of offsets creates a smoother and more realistic blurry output.}
    \label{fig:offset number}
\end{figure*}

\begin{table*}
\centering
\caption{Quantitative comparison of motion estimation on both synthetic and the GoPro \cite{nah2017deep} dataset.}
	\setcellgapes{2pt}\makegapedcells
	\begin{tabular}{c  c  c  c  c  c  c  c } 
		\toprule
		\multicolumn{2}{c}{Model}  & Sun \textit{et al.}\cite{sun2015learning} & Gong \textit{et al.} \cite{gong2017motion}  & Zero constraint (ZC) &  Linear & B-d Linear & Quadratic\\ 
		\midrule
		\multirow{3}{*}{Synthetic}& PSNR & 29.34 & 37.61 & 37.62 & 37.34  & 38.64 & \textbf{38.9}\\ 
		 & SSIM & 0.9001 & 0.9818 & 0.9763 & 0.9857 & 0.9872 & \textbf{0.9882}\\ 
		 & MSE  & 50.12 & 10.05 & - & 7.42  & 7.16 & \textbf{3.27} \\
		\midrule
		\multirow{2}{*}{GoPro} & PSNR & 29.68 & 30.61 & \textbf{35.82} & 33.45 & 33.79 & 34.68\\ 
		& SSIM & 0.9282 & 0.9363 & \textbf{0.9800} & 0.9669 & 0.9687 & 0.9740\\ 
		\midrule
		\multicolumn{2}{c}{Runtime(s)} & 45.2 & 8.4 &  0.011 & 0.011 & 0.011 & 0.011 \\
		\bottomrule
		\hline
	\end{tabular} 
	\label{tb:compare syn blurring}
	
\end{table*}

\begin{table*} 
    \begin{minipage}[t]{0.5\textwidth}
    \centering
    \makeatletter\def\@captype{table}\makeatother
    \caption{Comparison for the setting of offset numbers $N$.}
    \scalebox{0.96}{
    \begin{tabular}{ c c c c } 
		\toprule
		$\#$ of motion offsets & 5 & 9 & 15\\ 
		\midrule
		PSNR & 34.09 & 34.52 & \textbf{34.68} \\ 
		SSIM & 0.9668 & 0.9727 & \textbf{0.974}  \\
		\bottomrule
	
		\end{tabular} 
		\label{tb2:offset number}
    }
    
    \end{minipage} 
    \ \ \
    \begin{minipage}[t]{0.45\textwidth}
    \centering
    \makeatletter\def\@captype{table}\makeatother
    \caption{Ablation study for loss function.}
    \scalebox{0.96}{
    \begin{tabular}{ c c c c c } 
	\toprule
	 & Proposed & w/o SSIM &  w/o tv & w/o reg\\ 
	\midrule
	PSNR & \textbf{34.68} & 34.16 & 33.96 & 34.56\\ 
	SSIM & \textbf{0.974}& 0.97 & 0.9672 & 0.9727 \\
	\bottomrule
    \end{tabular} 
    \label{tb3:blurring loss ablation}
    }
    
    \end{minipage} \vspace{-2 mm}
\end{table*}

\subsection{Evaluation of Motion Offset Estimation}\label{sec:evaluation blurring}
We compare the proposed exposure trajectory recovery with one conventional blur kernel estimation method (Xu \textit{et al.} \cite{xu2013unnatural}) and two recent learning-based blur kernel estimation methods (Sun \textit{et al.} \cite{sun2015learning} and Gong \textit{et al.} \cite{gong2017motion}). Our comparisons are based on both the synthetic and GoPro datasets.

\begin{figure}[t]
    \centering
    \includegraphics[width=0.95\linewidth]{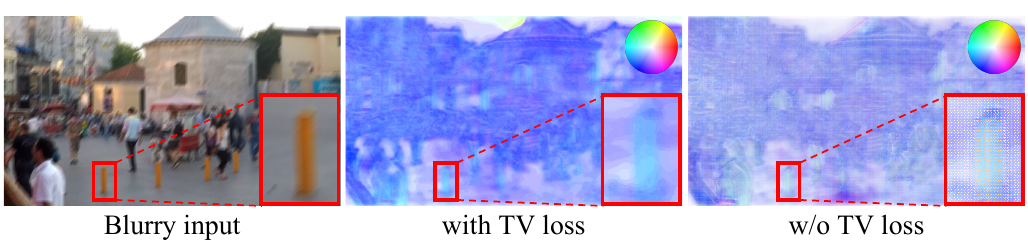} \vspace{-2 mm}
    \caption{\Revision{Visualize the magnitude and direction of motion offsets in a color coded map. Different colors represent different directions. The model with TV loss generates a much more smooth color map than the model w/o TV loss (best view in high resolutions).}}\vspace{-4 mm}
    \label{fig:vector direction}
\end{figure} 

\noindent \textbf{Evaluation Metrics.} In order to evaluate the accuracy of motion estimation, we calculate the PSNR and SSIM metrics between the input blurry image and reblurred image via estimated blur kernel/motion offsets for both datasets. Specifically, the reblurred results of \cite{sun2015learning} and \cite{gong2017motion} can be obtained by convolving a sharp image with the estimated motion flow map. We also apply the MSE metric of motion to evaluate the synthetic data. This metric defines the mean squared error between the ground truth motion and estimated motion~\cite{gong2017motion}. The MSE is easy to calculate in \cite{sun2015learning} and \cite{gong2017motion} since their estimated blur kernels share the same form as the ground truth, namely 2D vectors. However, our motion offsets are a set of points, so we calculate the vector of two endpoints as a simplification based on the assumption that the motion is linear. Note that we only provide the kernel visualization results of \cite{xu2013unnatural}, since its blur kernel cannot be represented as a pixel-wise motion flow map like the others.

	\begin{figure*} [ht]
		\centering
		\includegraphics[width=\linewidth]{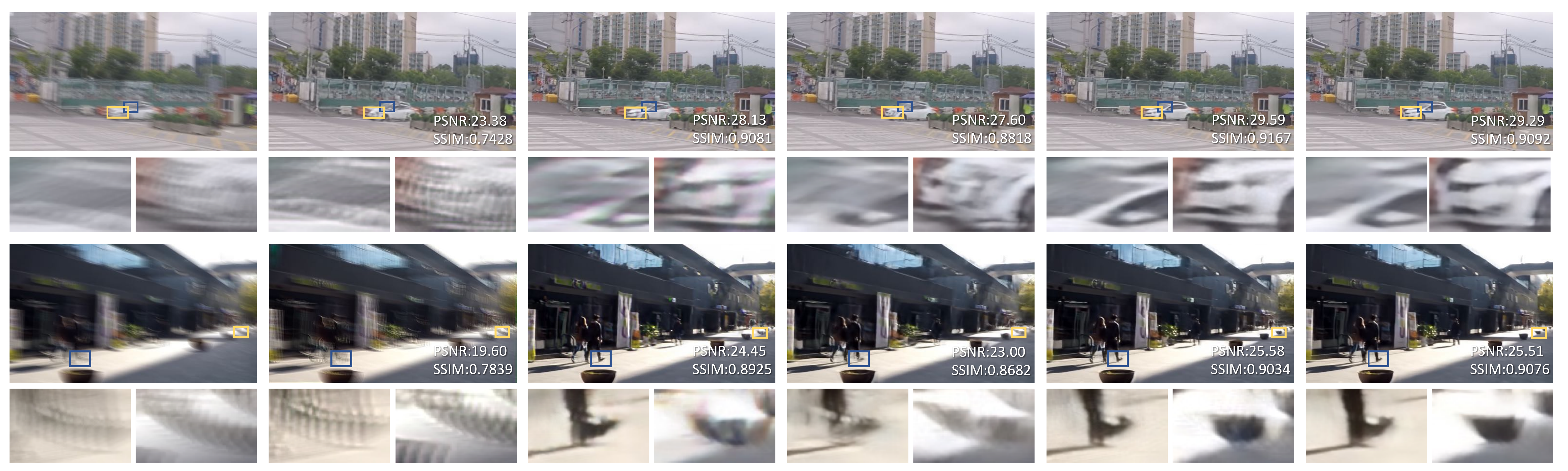}
		\vspace{-7mm}
		\caption{Visual comparison with GoPro dataset. From left to right, we show input, deblurring result of 
		\Revision{DeblurGAN-V2 \cite{kupyn2019deblurgan}}, 
		Gao \textit{et al.} 
		\cite{gao2019dynamic}, DMPHN~\cite{zhang2019deep}, ours, and stack(4)-DMPHN~\cite{zhang2019deep} (best view in high resolutions). }
		\label{Fig: compare gopro}
	\end{figure*}

	\begin{figure*}[t!]
    \centering
    \includegraphics[width=0.9\linewidth]{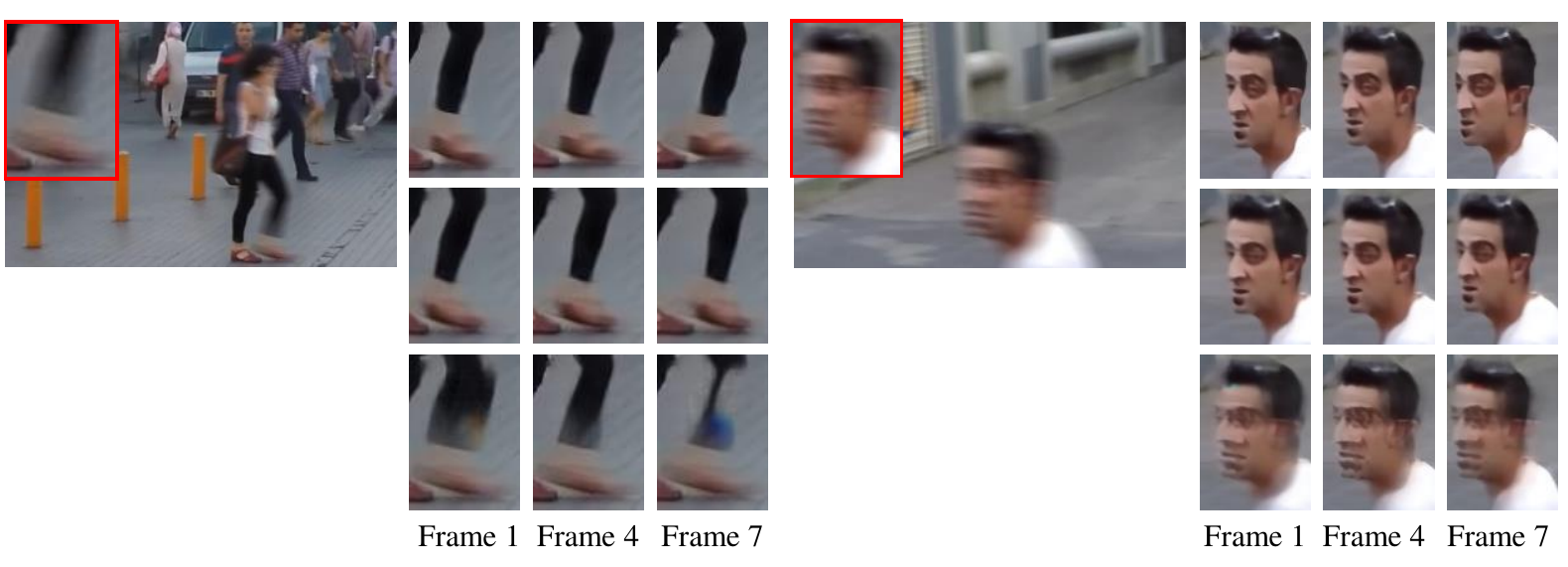}
    \vspace{-3.5 mm}
    \caption{Comparison of video extraction results. In the top-down order, we show ours, result of \cite{purohit2019bringing}, result of \cite{jin2018learning}.}
    \label{fig:video}
    \end{figure*}

\noindent \textbf{Motion Estimation on the Synthetic Dataset.}
Table~\ref{tb:compare syn blurring} shows our quantitative comparisons on the synthetic dataset. Our models with different constraints achieve comparable or better performance to \cite{gong2017motion}. It is noteworthy that \cite{gong2017motion} is learned in a supervised manner, yet our training scheme need no ground-truth motion as supervision. \Revision{Although the synthetic blur is linear, we can see the two non-linear constraint models (our b-d linear and quadratic) producing better reblurring PSNR results than the two linear constraint models (\cite{gong2017motion} and our linear), we infer that (i) comparing with linear model, the 
non-linear model has a higher degree of freedom brought by the estimation of two endpoints(only one for linear model), thus a greater fitting ability; (ii) the motion field of synthetic blur changes continuously and form into curves in the space (Fig. \ref{fig:motion_syn}). Overall, although the blurry kernel on each pixel is linear, the quadratic model has better representation ability and delivers a more accurate approximation for continuous change modeled in the synthetic dataset.}
    
We can also make some observations from Fig. \ref{fig:motion_syn}. Xu \textit{et al.} \cite{xu2013unnatural} generates non-trajectory kernels, for which we can only vaguely observe the flow after post-processing. Since Sun \textit{et al.} \cite{sun2015learning} performs a patch-level prediction from a blurry input, it is usually misled by the smooth area. Gong \textit{et al.} \cite{gong2017motion} shows more continuity across space, but there is also the possibility of when a region of predictions going wrong. Conversely, ours are more accurate and can perfectly fit into linear motion regardless of the employed constraints. 

    \begin{figure*}[ht]
    \centering
        \includegraphics[width=0.95\linewidth]{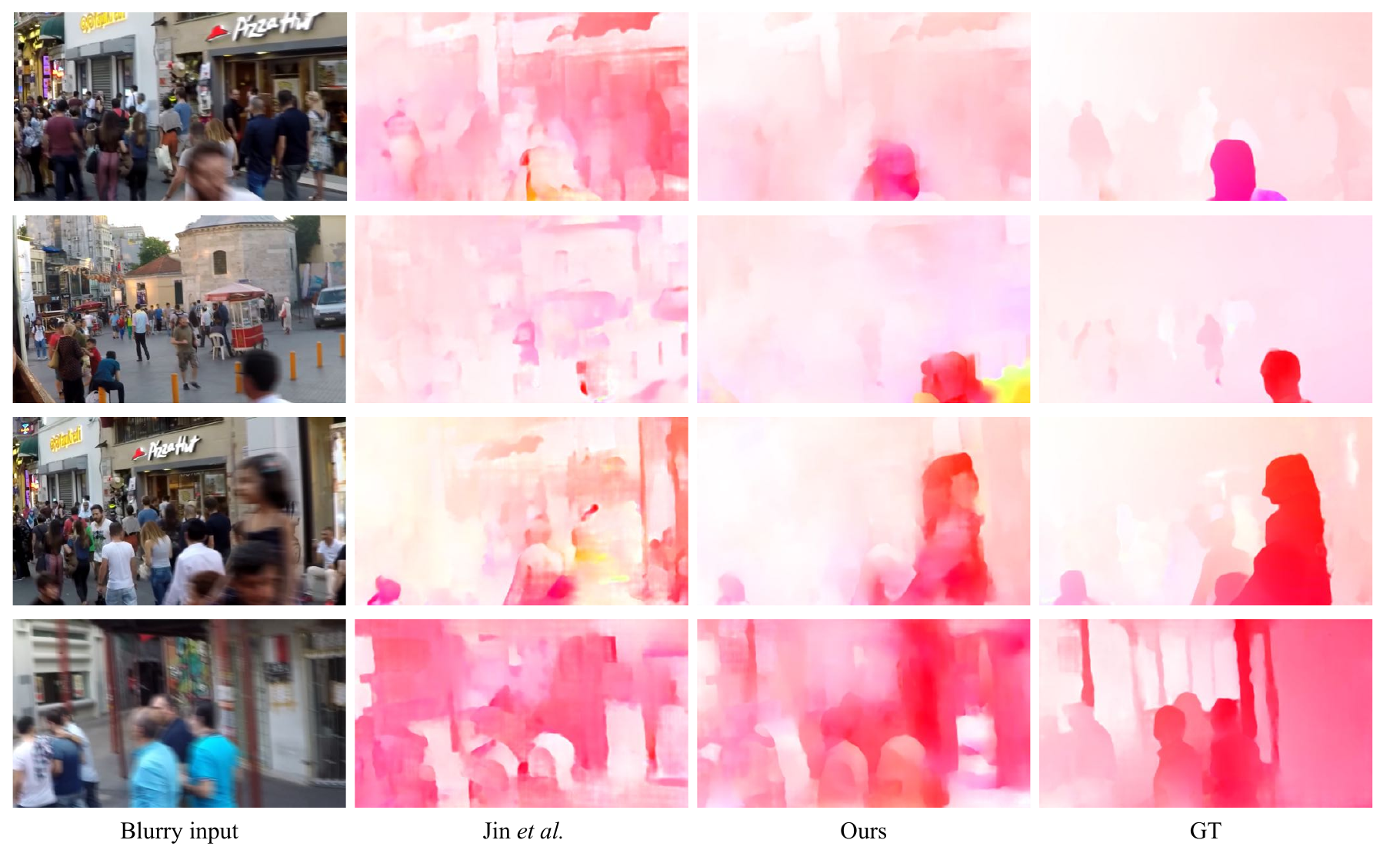}
         \vspace{-2.mm}
        \caption{\Revision{Comparison of optical flow from the extracted videos. The optical flow is calculated from the first and last frame of the extracted videos (Jin \textit{et al.}~\cite{jin2018learning} and ours) and the ground truth high-frame-rate video (GT).}}
        \label{fig:optical flow}
    \end{figure*}
    
    \begin{figure*}[ht]
    \centering
        \includegraphics[width=0.95\linewidth]{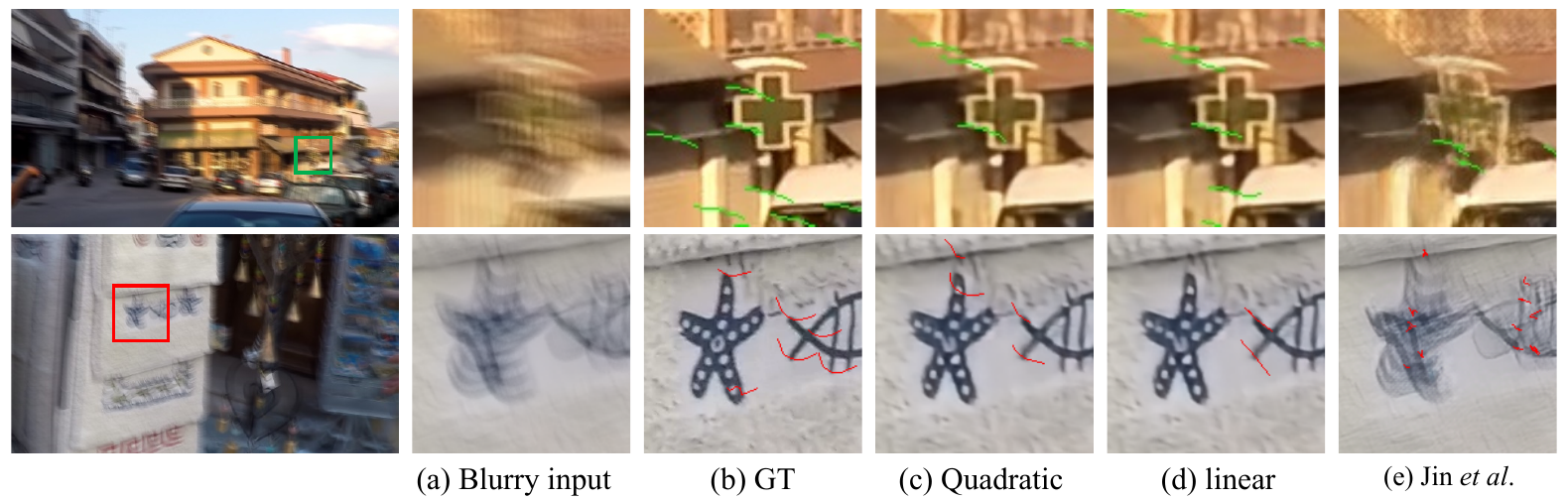}
         \vspace{-2.mm}
        \caption{Visualized trajectory result of extracted frames (best view in high resolutions).}
        \label{fig:visualization traj}

    \end{figure*}
    
    \begin{figure*}[ht]
    \centering
        \includegraphics[width=0.95\linewidth]{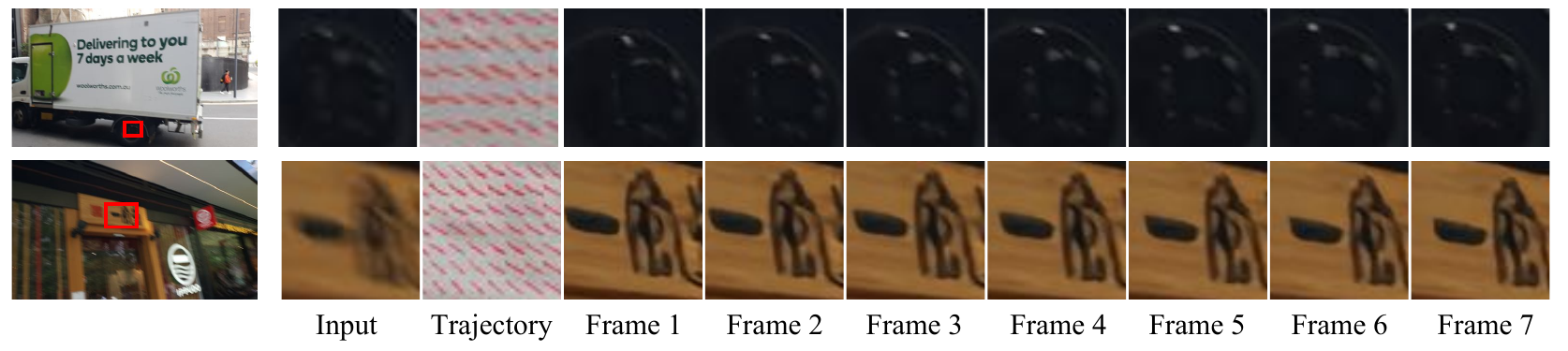}
        \vspace{-2 mm}
        \caption{Video extraction results with real images. More video results are provided in our supplementary video.}
        \label{fig:video extraction}
        \vspace{-3mm}

    \end{figure*}  
    
\noindent \textbf{Motion Estimation on the GoPro Dataset.} Since there is no motion ground truth for the GoPro dataset, the methods in \cite{sun2015learning,gong2017motion} cannot train their networks. Here, we employ their models pre-trained on the synthetic dataset and then test them on the GoPro test set. It is unfair to directly compare these results with our own; however, to our best knowledge, no other method is trained without motion ground truths, so their results seem to be a legitimate reference. 

Table~\ref{tb:compare syn blurring} shows that the performance of \cite{sun2015learning,gong2017motion} decreases significantly with more complex dynamic scenes. This decrease in quality can also be observed in the example in Fig. \ref{fig:motion_gopro} and Fig.~\ref{fig:motion_gopro_reblur}. \Revision{We first provide an example of visualized blur kernels estimated from different models in Fig.~\ref{fig:motion_gopro}}. As we can see, Xu \textit{et al.} \cite{xu2013unnatural} fails to estimate dynamic motion blur, and Sun \textit{et al.} \cite{sun2015learning} is obviously inaccurate and tends to generate spatially uniform kernels. The results using Gong \textit{et al.} \cite{gong2017motion}, although spatially variant, tend to produce many non-blurry regions. Our results, however, show a different flow direction in the background and foreground, \textit{e.g.,} the moving car. \Revision{To better demonstrate the accuracy and effectiveness of our proposed quadratic trajectory, we show more examples of the estimated motions and the corresponding reblurring regions in Fig.~\ref{fig:motion_gopro_reblur}. Our models generate more accurate reblurring results compared existing works. Also, the quadratic model is more effective in recovering the quadratic motion trajectory. Note that since the warping through motion offsets is conducted backward rather than forward, the exposure trajectory is center-symmetrical to the object moving trajectory. } 


\noindent \textbf{Ablation Studies.}
First, we discuss the setting of offset numbers $N$. As shown in Table~\ref{tb2:offset number}, the model with $N=15$ is notably better than the other models. The visual differences produced by altering the offset numbers are shown in Fig.~\ref{fig:offset number}. There is ghosting artifact with the model with 5 offsets, which becomes smoother as the offset number increases from 5 to 15. With 15 offsets, the result is very close to the ground truth image. Since increasing the number of offsets has little effect on performance, we set $N=15$, considering the balance between performance and efficiency.

\begin{table*}[ht!]
\begin{center}
\scriptsize
\setcellgapes{2pt}\makegapedcells
\caption{Quantitative deblurring results on GoPro dataset.}\vspace{-2mm}
	\begin{tabular}{c c c c c c c c c c} 
		\toprule
		Model &  Gong \cite{gong2017motion} & Nah \cite{nah2017deep} &  Tao \cite{tao2018scale} & \Revision{DeblurGAN~\cite{kupyn2018deblurgan}} & \Revision{DeblurGAN-V2 \cite{kupyn2019deblurgan}} & Gao \cite{gao2019dynamic} & DMPHN\cite{zhang2019deep}  & Ours & Stack(4)-DMPHN\cite{zhang2019deep}\\
		\midrule
		PSNR & 26.89 & 29.08 & 30.26 & 28.7 & 29.55 & 30.92 & 30.21 & \underline{31.05} & \textbf{31.20} \\ 
		SSIM & 0.8639 & 0.9135 & 0.9342 & 0.9270 & 0.9340 & 0.9421 & 0.9345 & \textbf{0.9485} & \underline{0.9453} \\
		Size(MB) & 54.1 & 303.6 & 33.6 & 35.4 & 244.5 & 46.5 & \textbf{21.7} & \underline{26.3} & 86.8  \\
		\bottomrule
	\end{tabular}
    \vspace{-2mm}
    \label{table:deblur-results}
\end{center}
\end{table*}

\begin{table*}[ht!]
\begin{center}
\setcellgapes{2pt}\makegapedcells
\caption{\Revision{Ablation study of motion-aware convolution on GoPro dataset.}}\vspace{-2mm}
	\begin{tabular}{ c c c c c c} 
		\toprule
		Model &  \Revision{Baseline (w/o MA Conv.)} & Zero constraint (ZC) & Linear &  B-d Linear & Quadratic  \\
		\midrule
		PSNR & 30.21 & 30.79 & 30.82 & 31.04 & \textbf{31.05} \\ 
		SSIM & 0.9345 & 0.9459 & 0.9462 & 0.9483 & \textbf{0.9485} \\
		\bottomrule
	\end{tabular}
    \label{table:deblur-ablation}
\end{center}
\end{table*}

To demonstrate the effectiveness of the proposed loss function, we trained a model with all the proposed losses (Model \textbf{Proposed}), one without the SSIM loss (Model \textbf{w/o SSIM}), one without the total variation loss (Model \textbf{w/o tv}), and one without the regulation loss (Model \textbf{w/o reg}). The quantitative results are shown in Table~\ref{tb3:blurring loss ablation}. The proposed loss combination is better than those without certain losses. 
\Revision{The total variation loss which encourage the local uniformity not only improves the motion estimation accuracy, but also solves the ambiguity of the motion direction. The ambiguity of direction always exists since a single blurry image barely contains the direction information, yet the predicted directions under total variance loss will be spatially continuous rather than random. as shown in Fig.~\ref{fig:vector direction}, we visualize the magnitude and direction of optimized motion offsets by calculating the vector subtraction between the first motion offset and the last motion offset. The model with TV loss generates a smooth color map, while the model w/o TV loss generates the color map with noise dots, which means the directions of these pixels are discontinuous or even opposite to the adjacent pixels.} Moreover, although the regulation loss has little influence on the metrics, it prevents the network from estimating large offsets in the smooth area.
    
\subsection{Evaluation of Dynamic Scene Deblurring}
We quantitatively and qualitatively compare our method with recent state-of-the-art dynamic scene deblurring methods: \Revision{DeblurGAN(-v2) \cite{kupyn2018deblurgan,kupyn2019deblurgan} based on a conditional GAN to obtain a more realistic texture;} Nah \textit{et al.}~\cite{nah2017deep}, Tao \textit{et al.} \cite{tao2018scale}, and Gao \textit{et al.} \cite{gao2019dynamic} built multi-scale networks but with different parameter sharing and parameter independence schemes; Zhang \textit{et al.}~\cite{zhang2019deep} applies a Deep Multi-Patch Hierarchical Network (DMPHN), which is also the backbone network of our method. We also provide the deblurring results with \cite{gong2017motion} as representative of conventional MAP optimization. The quantitative results are presented in Table~\ref{table:deblur-results}.

As illustrated in Table~\ref{table:deblur-results}, our motion-aware deblurring network achieves comparable results to current state-of-the-art methods with respect to PSNR and achieved slightly better result with respect to SSIM. Note that, our model achieves such performance using a single-stack, which only costs about 30\% of the model size compared to the model of the stack(4)-DMPHN. Considering the only difference between our model with DMPHN is the proposed motion-aware convolutional layer, it contributes 0.84 and 0.014 increasing in PSNR and SSIM, respectively. Also, as shown in Fig.~\ref{Fig: compare gopro}, the visual results of ours are almost the same as stack(4)-DMPHN, while better than the other methods.

Besides verifying that the learned exposure trajectories could contribute to dynamic scene deblurring, we also conduct ablation studies to discuss the effects of different kinds of exposure trajectories. As Table~\ref{table:deblur-ablation} shows, compared to the baseline model, all kinds of exposure trajectories could improve the deblurring performance. The Model Linear and B-d linear perform slightly inferior to the Model Quadratic, owning to the less accurate of the motion estimation. Note that, though the exposure trajectory learned with zero-constraint (ZC) achieves the best score in above reblurring experiments (Table~\ref{tb:compare syn blurring}-GoPro), it demonstrates less effect in our deblurring module. A reasonable explanation is that the zero-constraint motion offsets are only one of the ill-posed solutions for reblurring reconstruction, yet it will not be the most accurate trajectory estimation.

\subsection{Evaluation of Video Extraction}

To evaluate the performance of our approach for video extraction, we compare our results with those of Jin \textit{et al.} \cite{jin2018learning}, Purohit \textit{et al.} \cite{purohit2019bringing} and \Revision{Zhang \textit{et al.} \cite{zhang2020every}}. 
\Revision{Since the source codes of methods of Purohit \textit{et al.} \cite{purohit2019bringing} and Zhang \textit{et al.} \cite{zhang2020every} are not public yet, we can only provide the data recorded in their papers. We first directly compare the accuracy of the extracted frames. In Table \ref{table:video extraction}, the PSNR and SSIM metrics are applied to measure the deblurring performance of the centered frame, and our method achieves the highest scores. Also, we can see from the Fig.~\ref{fig:video} that the deblurring result (frame 4) and video extraction results (frames 1 and 7) of \cite{jin2018learning} both perform poorly when handling large blur.} The results of \cite{purohit2019bringing} and our own show relatively sharp frames and clear object movements. 

\begin{table}[ht!]
\begin{center}
\caption{\Revision{Quantitative comparison of video extraction on GoPro dataset.}}\vspace{-2mm}
	\begin{tabular}{ l c c c} 
		\toprule
		Method \ \ \ \  & \ \ \ PSNR \ \ \ & \ \ \ SSIM \ \ \  & \ \ \ EPE  \ \ \  \\
		\midrule
		Jin \textit{et al.}~\cite{jin2018learning} & 26.98 & 0.881 & 9.32 \\ 
		Purohit \textit{et al.}~\cite{purohit2019bringing} & 30.58 & 0.941 & - \\
		Zhang \textit{et al.}~\cite{zhang2020every} & 30.64 & 0.942 & 10.03 \\
		Jin \textit{et al.} + Ours & 26.98 & 0.881 & \textbf{6.07} \\
		Ours & \textbf{31.05} & \textbf{0.949} & 6.09 \\
		\bottomrule
	\end{tabular}
    \vspace{-3mm}
    \label{table:video extraction}
\end{center}
\end{table}

\Revision{Then we further validate the accuracy of the motion in extracted videos through comparisons on the optical flow estimated from the synthesized videos. We follow the approach of calculating end-point error (EPE) in \cite{zhang2020every}. Specifically, we first estimate optical flow from the first generated frame to the last one with PWC-net \cite{sun2018pwc} and vice versa. Then, EPE calculates the errors between estimated flows and the flow estimated from ground truth high-frame-rate video. The lower EPE value is chosen as the result, since the direction is uncertain. As shown in Table ~\ref{table:video extraction}, our methods achieve the best EPE. For a fair comparison, we combine the centered frame generated from model of Jin \textit{et al.} \cite{jin2018learning} with our motion offsets to warp the other frames, termed Jin \textit{et al.} \cite{jin2018learning} + Ours, demonstrating that our improvement on EPE score comes mainly from the more accurate motion estimation rather than a superior deblurring result. Due to the limitation of using optical flow to evaluate video clarity, we obtain similar EPE scores from the generated video of Jin \textit{et al.} \cite{jin2018learning} + ours and ours. To summary, the metrics in Table ~\ref{table:video extraction} demonstrate that the extracted video from our model is sharper and the encoded motion is more accurate. Qualitative comparison of the visualized optical flow in Fig.~\ref{fig:optical flow} also shows that the videos extracted using our method present a more accurate optical flow compared to \cite{jin2018learning}. 
} 

\Revision{Besides the validation of optical flow from first frame to last frame, we also validate the effectiveness of our quadratic trajectory.} As shown in Fig. \ref{fig:visualization traj}, we visualize the trajectory using feature point tracking \cite{shi1994good}. Our quadratic motion offsets better fit the curve to the ground truth, especially in the second example. 

\Revision{Our exposure trajectory recovery framework employs blurry/sharp image pairs as training data. It delivers impressive optical flow estimation/trajectory results of dynamic scenes without accessing any motion supervision. Moreover, our motion offsets can be combined with any state-of-the-art deblurring method to generate even sharper video clips, while other methods need to train a whole new model. Finally, our model can generate arbitrary numbers of frames, while \cite{jin2018learning} can only achieve a fixed number. Although \cite{purohit2019bringing,zhang2020every} can also generate slow-motion videos, they need to conduct a iterative generation which increase the inference time. However, we only need to interpolate in our trajectory after a single forward prediction.} As a result, our network is more compact and faster. The runtimes of \cite{jin2018learning}, \cite{purohit2019bringing}, and our model are 1.1 s, 0.39 s, and 0.22 s respectively. We also provide an example of video extraction from real images in Fig.~\ref{fig:video extraction}, demonstrating a good generalization ability of our proposed method. More video results can be found in our supplementary video. 

\section{\Revision{Limitations}}\label{Sec:limitations}
\begin{figure}
    \centering
    \includegraphics{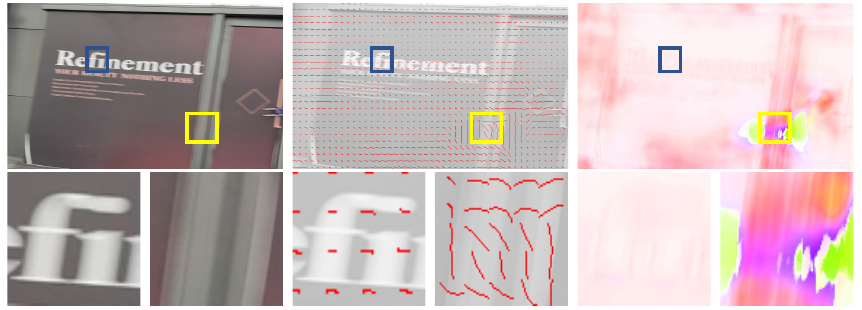}
    \caption{\Revision{A failure case on real-world blurry image. From left to right are the blurry input, the visualized exposure trajectory and color-coded motion offset map. The obvious error exists where the estimated motion is very small or has a wrong direction.}}
    \label{fig:failure}
\end{figure}

There are two main limitations that existed in our proposed method. First, a more complicated motion may be caused by a large camera shake or a highly dynamic scene, which may need to be modeled by a higher order of exposure trajectory. In these situations, our quadratic constrained exposure trajectory can only act as an approximation of the ground-truth trajectory. However, since the motion captured in the dataset is relatively small, these situations rarely happen in the existing blurry image datasets. Second, similar to most learning-based deep models, our method may have generalization issues, it may fail to handle blur patterns that have a domain gap with the blur in training data. Considering our model still need to be trained on blurry/sharp image pairs, \textit{i.e.,} the GoPro dataset, which is an approximation of real-world blurry images, our well-trained network may fail in recovering the trajectory when encounter unseen real-world blur (Fig.~\ref{fig:failure}). Solving this domain-gap problem is very challenging since an unsupervised training strategy is required for the unpaired real-world data. However, we believe our fully differentiable (re)blurring module can potentially contribute to unsupervised motion estimation/image deblurring, since it can provide a cycle-consistency loss. In the future, we will continue to devote to develop fully unsupervised deblurring and motion estimation methods for motion estimation.

\section{Conclusion}\label{Sec:Conclusion}
Here we propose an exposure trajectory recovery scheme to generate motion offsets which are superior to conventional blur kernels in many respects. By imposing different constraints, these offsets can fit into different exposure trajectories. Moreover, we utilize the learned motion offsets for image deblurring and video extraction from a single blurry image. Experiments show that our motion offsets can produce useful information for solving these tasks. However, the learned exposure trajectories are still limited to motion of constant acceleration, and may not perfectly fit real situations. We will further devote to provide more accurate motion estimation and further improve the deblurring and video extraction tasks.


%

\ifCLASSOPTIONcompsoc
\section*{Acknowledgments}
The authors would like to thank anonymous reviewers and the handlining editor for their constructive comments. This work was supported by Australian Research Council Projects FL-170100117, IH-180100002, IC-190100031.

\ifCLASSOPTIONcaptionsoff
  \newpage
\fi



\bibliographystyle{IEEEtran}
\bibliography{egbib_abbv}
%



%

\begin{IEEEbiography}[{\includegraphics[width=1in,height=1.25in,clip,keepaspectratio]{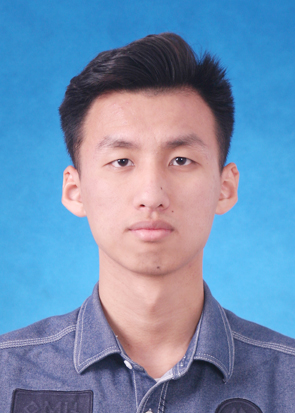}}]{Youjian Zhang} received the BE degree in electronics science and technology from Shanghai Jiao Tong University, Shanghai, China, in 2018. He is currently working toward the PhD degree in computer science from the University of Sydney, Camperdown, NSW, Australia. His research interests are computer vision with emphases on image deblurring, image/video quality enhancement and restoration. Recently, one of his works has been accepted by the Conference on Neural Information Processing Systems (NeurIPS).
\end{IEEEbiography}

\begin{IEEEbiography}[{\includegraphics[width=1in,height=1.25in,clip,keepaspectratio]{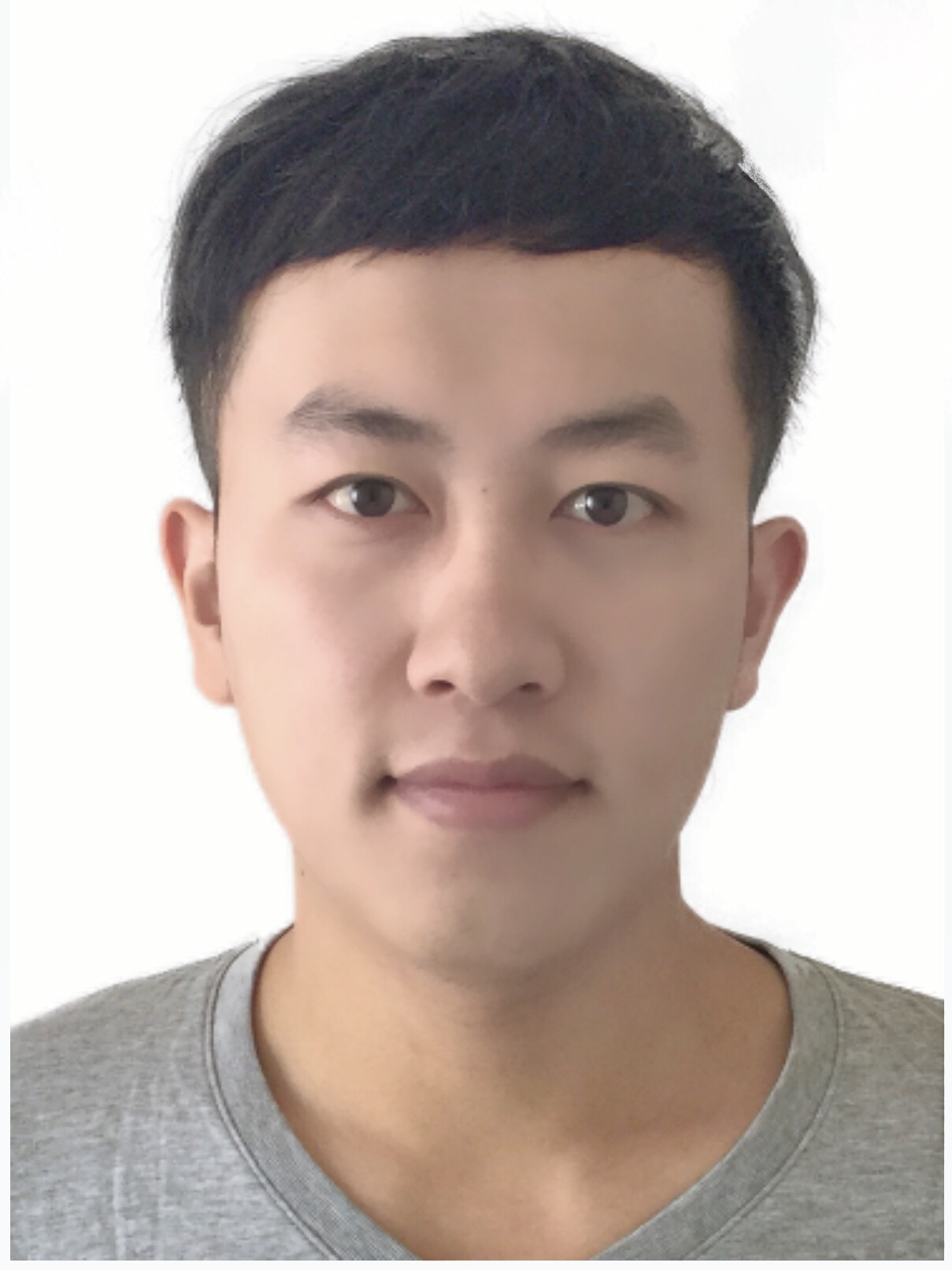}}]{Chaoyue Wang} is a postdoctoral researcher in Machine Learning and Computer Vision at the School of Computer Science, The University of Sydney. He received a bachelor degree from Tianjin University (TJU), China, and a Ph.D. degree from the University of Technology Sydney (UTS), Australia. His research outcomes have been published in prestigious journals and prominent conferences, such as IEEE T-EVC, IEEE T-IP, NeurIPS, CVPR, IJCAI. He received the Distinguished Student Paper Award in the 2017 International Joint Conference on Artificial Intelligence (IJCAI-17).
\end{IEEEbiography}

\begin{IEEEbiography}[{\includegraphics[width=1in,height=1.25in,clip,keepaspectratio]{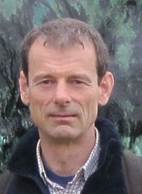}}]{Stephen J. Maybank} (Fellow, IEEE) received the BA degree in mathematics from King’s CollegeCambridge, in 1976 and the PhD degree in computer science from Birkbeck College, University of London, in 1988. He is currently a professor with the School of Computer Science and Information Systems, Birkbeck College. His research interests include the geometry of multiple images, camera calibration, visual surveillance, \emph{etc}. He is a fellow of the IEEE and fellow of the Royal Statistical Society.
\end{IEEEbiography}

\begin{IEEEbiography}[{\includegraphics[width=1in,height=1.25in,clip,keepaspectratio]{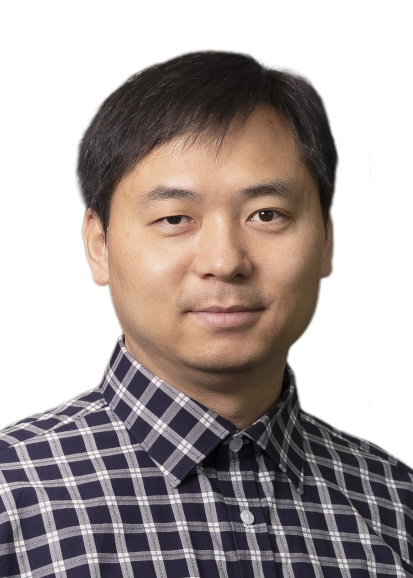}}]{Dacheng Tao} (Fellow, IEEE) is an advisor and chief scientist of the digital science institute in the University of Sydney. He mainly applies statistics and mathematics to artificial intelligence and data science, and his research is detailed in one monograph and over 200 publications in prestigious journals and proceedings at leading conferences. He received the 2015 and 2020 Australian Eureka Prize, the 2018 IEEE ICDM Research Contributions Award, and the 2021 IEEE Computer Society McCluskey Technical Achievement Award. He is a Fellow of the Australian Academy of Science, AAAS, ACM and IEEE.
\end{IEEEbiography}









\end{document}